\documentclass[nohyperref]{article}

\usepackage{microtype}
\usepackage{graphicx}
\usepackage{subfigure}
\usepackage{booktabs} % for professional tables
\usepackage{wrapfig}

\usepackage{epsfig}
\usepackage{bbm}
\usepackage{xcolor}
\usepackage{enumitem}
\usepackage{multirow}
\definecolor{darkgreen}{rgb}{0.0, 0.5, 0.0}

\newcommand{\specialcell}[2][c]{%
 \begin{tabular}[#1]{@{}c@{}}#2\end{tabular}}

\usepackage{hyperref}

\usepackage[accepted]{icml2022}

\usepackage{amsmath}
\usepackage{amssymb}
\usepackage{mathtools}
\usepackage{amsthm}

\usepackage[capitalize,noabbrev]{cleveref}

\theoremstyle{plain}
\newtheorem{theorem}{Theorem}[section]
\newtheorem{proposition}[theorem]{Proposition}

\theoremstyle{definition}
\newtheorem{definition}[theorem]{Definition}

\theoremstyle{remark}

\usepackage[textsize=tiny]{todonotes}

\icmltitlerunning{DOC$^3$}

\begin{document}

\twocolumn[
\icmltitle{DOC$^3$ -  Deep One Class Classification using Contradictions}

% It is OKAY to include author information, even for blind
% submissions: the style file will automatically remove it for you
% unless you've provided the [accepted] option to the icml2022
% package.

% List of affiliations: The first argument should be a (short)
% identifier you will use later to specify author affiliations
% Academic affiliations should list Department, University, City, Region, Country
% Industry affiliations should list Company, City, Region, Country

% You can specify symbols, otherwise they are numbered in order.
% Ideally, you should not use this facility. Affiliations will be numbered
% in order of appearance and this is the preferred way.
\icmlsetsymbol{equal}{*}

\begin{icmlauthorlist}
\icmlauthor{Sauptik Dhar}{to}
\icmlauthor{Bernardo Gonzalez Torres}{goo}
\end{icmlauthorlist}

\icmlaffiliation{to}{Independent, Santa Clara, CA, USA}
\icmlaffiliation{goo}{Intuition Machines, Inc., San Francisco, CA, USA}

\icmlcorrespondingauthor{Sauptik Dhar}{sauptik.dhar@gmail.com}

% You may provide any keywords that you
% find helpful for describing your paper; these are used to populate
% the "keywords" metadata in the PDF but will not be shown in the document
\icmlkeywords{Anomaly Detection, Deep Learning, Learning from Contradictions}

\vskip 0.3in
]

% this must go after the closing bracket ] following \twocolumn[ ...

% This command actually creates the footnote in the first column
% listing the affiliations and the copyright notice.
% The command takes one argument, which is text to display at the start of the footnote.
% The \icmlEqualContribution command is standard text for equal contribution.
% Remove it (just {}) if you do not need this facility.

%\printAffiliationsAndNotice{}  % leave blank if no need to mention equal contribution
\printAffiliationsAndNotice{} % otherwise use the standard text.

\begin{abstract}
This paper introduces the notion of learning from contradictions (a.k.a Universum learning) for deep one class classification problems. We formalize this notion for the widely adopted one class large-margin loss \cite{scholkopf2001estimating}, and propose the Deep One Class Classification using Contradictions (DOC$^3$) algorithm. We show that learning from contradictions incurs lower generalization error by comparing the Empirical 
Rademacher Complexity (ERC) of DOC$^3$ against its traditional inductive learning counterpart. Our empirical results demonstrate the efficacy of DOC$^3$ compared to popular baseline algorithms on several real-life data sets.
\end{abstract}

\section{Introduction} \label{sect:intro}
Anomaly Detection (AD) is one of the most widely researched problem in the machine learning community \cite{chandola2009anomaly}. In its basic form, the task of Anomaly Detection (AD) involves discerning patterns in data that do not conform to expected `normal' behavior. These non-conforming patterns are referred to
as anomalies or outliers. Anomaly detection problems manifest in several forms in real-life like, defect detection in manufacturing lines, intrusion detection for cyber security, or pathology detection for medical diagnosis etc. There are several mechanisms to handle anomaly detection problems viz., parametric or non-parametric statistical modeling, spectral based, or classification based modeling \cite{chandola2009anomaly}. Of these, the classification based approach has been widely adopted in literature \cite{scholkopf2002learning,tax2004support,tan2016introduction, cherkassky2007learning}. One specific classification based formulation which has gained huge adoption is one class classification \cite{scholkopf2002learning,tax2004support}, where we design a parametric model to estimate the support of the `normal' class distribution. The estimated model is then used to detect `unseen' abnormal samples. 

With the recent success of deep learning based approaches for different machine learning problems, there has been a surge in research adopting deep learning for one class problems \cite{ruff2021unifying,pang2020deep,chalapathy2019deep}. However, most of these works adopt an inductive learning setting. This makes the underlying model estimation data hungry, and perform poorly for applications with limited training data availability, like medical diagnosis, industrial defect detection, etc. The \textit{learning from contradictions} paradigm (popularly known as Universum learning) has shown to be particularly effective for problems with limited training data availability \cite{vapnik06,sinz08,weston06,chen09,cherkassky2011practical,shen12,dhar15,zhang17,xiao2021new}. However, it has been mostly limited to binary or multi class problems. In this paradigm, along with the labeled training data we are also given a set of unlabeled contradictory (a.k.a universum) samples. These universum samples belong to the same application domain as the training data, but are known not to belong to any of the classes. The rationale behind this setting comes from the fact that even though obtaining labels is very difficult, obtaining such additional unlabeled samples is relatively easier. These unlabeled universum samples act as \textit{contradictions} and should not be explained by the estimated decision rule. Adopting this to one class problems is not straight forward.  A major conceptual problem is that, one class model estimation represents unsupervised learning, where the notion of contradiction needs to be redefined properly. In this paper,
\begin{enumerate}[nosep]
\item \textbf{Definition} We introduce the notion of `Learning from contradictions' for one class problems (Definition \ref{def_univsetting}).
\item \textbf{Formulation} We analyze the popular one class hinge loss \cite{scholkopf2001estimating}, and extend it under universum settings to propose the Deep One Class Classification using contradictions DOC$^3$ algorithm.
\item \textbf{Generalization Error} We analyze the generalization performance of one class formulations under inductive and universum settings using Rademacher complexity based bounds, and show that learning under the universum setting can provide improved generalization compared to its inductive counterpart.
\item \textbf{Empirical Results} Finally, we provide an exhaustive set of empirical results in support of our formulation.    
\end{enumerate}

\section{One class learning under inductive  settings} \label{sect:inductive}
First we introduce the widely adopted inductive learning setting used for one class problems \cite{scholkopf2002learning,cherkassky2007learning}.

\begin{definition} {\normalfont (Inductive Setting)} \label{def_inductive}
Given i.i.d training samples from a single class $\mathcal{T}=(\mathbf{x}_i, \; y_i = +1)_{i=1}^n \sim \mathcal{D}_{\mathcal{X}|\mathcal{Y} = +1}^n$, with $\mathbf{x} \in \mathcal{X} \subseteq \Re^d$ and $y \in \mathcal{Y} = \{-1,+1 \}$; estimate a hypothesis $h^*:\mathcal{X} \rightarrow \mathcal{Y}$ from an hypothesis class $\mathcal{H}$ which minimizes, 
\begin{flalign} \label{eq_inductive}
 \underset{h \in \mathcal{H}}{\text{inf}} \; \mathbb{E}_{\mathcal{D}_{\mathcal{T}}}[\mathbbm{1}_{y \neq h(\mathbf{x})}]
\end{flalign}
$\mathcal{D}_{\mathcal{T}}$ is the training distribution (consisting of both classes) \\ 
$\mathcal{D}_{\mathcal{X}|\mathcal{Y} = +1}$ is class conditional distribution\\
$\mathbbm{1}(\cdot)$ is the indicator function, and \\ 
$\mathbb{E}_{\mathcal{D}_{\mathcal{T}}}(\cdot)$ is the  expectation under training distribution.
\end{definition} 
\noindent Note that, the underlying data generation process assumes a two class problem; of which the samples from only one class is available during training. The overall goal is to estimate a model which minimizes the error on the future test data, containing samples from both normal ($y = +1$) and abnormal classes ($y = -1$). Typical examples include, AI driven visual inspection of product defects in a manufacturing line; where images or videos of non-defective products are available in abundance. The goal is to detect `defective' (abnormal / anomalous) products through visual inspection in manufacturing lines \cite{bergmann2019mvtec,weimer2016design}. A popular loss function used in such settings is the  $\nu$-SVM loss \cite{scholkopf2001estimating},
\begin{flalign} \label{eq_nuSVM}
& \underset{\mathbf{w},\boldsymbol \xi, \rho}{\text{min}}\quad \frac{1}{2}||\mathbf{w}||_2^2 \;  + \;  \frac{1}{\nu n} \sum_{i=1}^n \xi_i -\rho && \\
&\text{s.t.} \quad \mathbf{w}^\top \phi(\mathbf{x}_i) \geq \rho -\xi_i, \quad \xi_i \geq 0; \quad \forall\; i = 1 \ldots n && \nonumber
\end{flalign}
\noindent where, $\nu \in (0,1]$ is a user-defined parameter which controls the margin errors $\sum_i \xi_i$ and the size of geometric $\frac{1}{||\mathbf{w}||}$ and functional $\rho$ margins. $\phi(\cdot): \mathcal{X} \rightarrow \mathcal{G}$ is a feature map. Typical examples include an \textit{empirical} kernel map (see Definition 2.15 \cite{scholkopf2002learning}) or a map induced by a deep learning network \cite{goodfellow2016deep}. The final decision function is given as, \; $h(\mathbf{x}) = \left\{
\begin{array}{l l}
    +1;\quad  \text{if} \; \mathbf{w}^\top \phi(\mathbf{x}_i) \geq \rho  \\
    -1;\quad  \text{else}
\end{array}\right. $. Note that, recent works like \cite{ruff2018deep} extend a different loss function which uses a ball to explain the support of the data distribution following \cite{tax2004support}. As discussed in \cite{scholkopf2001estimating}, most of the time these two formulations yield equivalent decision functions. For example, with kernel machines $\mathbf{K}(\mathbf{x},\mathbf{x}^{\prime}) = \phi(\mathbf{x})^\top \phi(\mathbf{x}^\prime)$ depending solely on $\mathbf{x} - \mathbf{x}^{\prime}$ (like RBF kernels), these two formulations are the same. Hence, most of the improvements discussed in this work translates to such alternate formulations. In this paper however, we solve the following one class Hinge Loss,
\begin{flalign} \label{eq_SVM}
& \underset{\mathbf{w}}{\text{min}}\quad \frac{1}{2}||\mathbf{w}||_2^2 \;  + \;  C \; L_T(\mathbf{w},\{\phi(\mathbf{x}_i)\}_{i=1}^n) && \\
& L_T(\mathbf{w},\{\phi(\mathbf{x}_i)\}_{i=1}^n) = \sum_{i=1}^n [1-\mathbf{w}^\top\phi(\mathbf{x}_i)]_+ \; ; \; [x]_+ = \text{max}(0,x) && \nonumber
\end{flalign}
to estimate the the decision function $f(\mathbf{x}) = \mathbf{w}^\top \phi(\mathbf{x}_i)$ and use the decision rule,
$h(\mathbf{x}) = \left\{
\begin{array}{l l}
    +1;\quad  \text{if} \; f(\mathbf{x}) \geq 1  \\
    -1;\quad  \text{else}
\end{array}\right.$. Here, the user-defined parameter $C$ controls the trade-off between explaining the training samples (through small margin error $\sum_{i=1}^n \xi_i $), and the margin size (through $||\mathbf{w}||_2^2$), which in turn controls the generalization error. For deep learning architectures we optimize using all the model parameters and equivalently regularize the entire matrix norm $||\mathbf{W}||_F^2$, see \cite{goyal2020drocc,ruff2018deep}. Note that, we solve one class Hinge loss \eqref{eq_SVM} for the two main reasons,
\begin{itemize}[nosep]
    \item[--] First, it has the advantage that $L_T(\mathbf{w},\phi(\{\mathbf{x}\}_{i=1}^n)) = \sum_{i=1}^n [1-\mathbf{w}^T\phi(\mathbf{x}_i)]_+ $ exhibits the same form as the traditional hinge loss used for binary classification problems \cite{vapnik06} and can be easily solved using existing  software packages \cite{pytorch, abadi2016tensorflow,pedregosa2011scikit}. Throughout the paper we refer \eqref{eq_SVM} using underlying deep architectures as Deep One Class DOC (Hinge) formulation.
    \item[--] Second, solving \eqref{eq_SVM} also provides the solution for  \eqref{eq_nuSVM}. This connection follows from Proposition \ref{prop_equivalence}.
\end{itemize}  
\begin{proposition} \label{prop_equivalence} Connection between \eqref{eq_nuSVM} and \eqref{eq_SVM}
\begin{enumerate}[label=\roman*]
    \item Any solution $\mathbf{w}$ of \eqref{eq_SVM} also solves \eqref{eq_nuSVM} with  $\nu = \frac{1}{Cn \delta}$; where $\delta > 0$ is a scalar that depends on the solution of \eqref{eq_SVM}. Further, this solution $(\mathbf{\hat w}, \rho)$ of \eqref{eq_nuSVM} is given as $\mathbf{\hat w} = \mathbf{w}\delta , \rho = \delta$.
    \item The decision function obtained through solving \eqref{eq_SVM} i.e., $\mathbf{w}^\top \phi(\mathbf{x}) - 1 = 0$ coincides with the decision function $\mathbf{\hat w}^\top \phi(\mathbf{x}) - \rho = 0$ obtained by solving \eqref{eq_nuSVM} using (i).
\end{enumerate}
\end{proposition}
\noindent All proofs are provided in Appendix.

\section{One class learning using Contradictions a.k.a Universum Learning} \label{sect:univ}
\subsection{Problem Formulation} \label{sect:formulation}

\begin{figure} 
\centering
\includegraphics[width=0.45\textwidth]{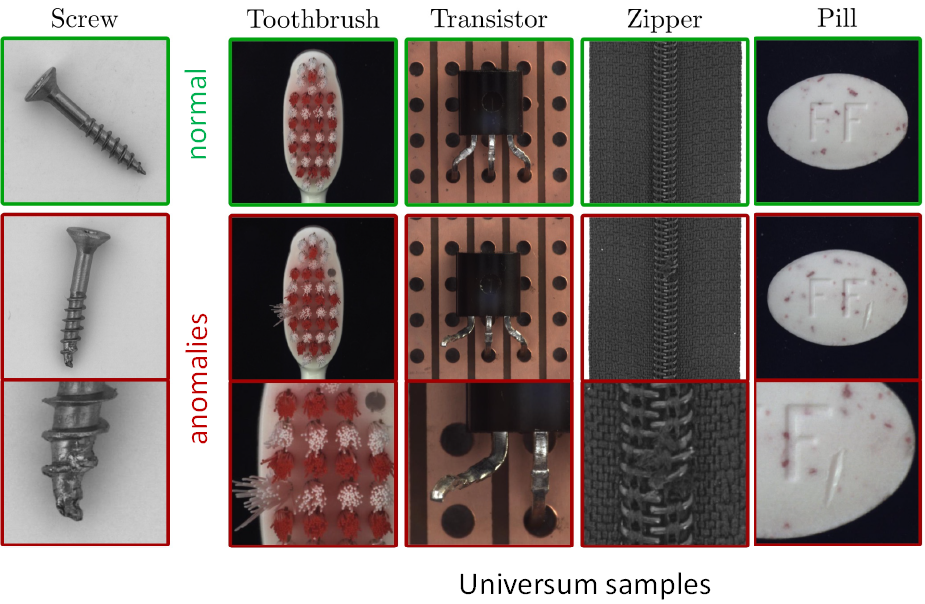} 
\vskip -0.1in
\caption{Visual inspection of anomalous screws in a manufacturing line \cite{bergmann2019mvtec}. Images of the other products act as universum samples. Such images are neither \textcolor{darkgreen}{normal} - screw nor \textcolor{red}{anomalous}-screw images and act as contradictions.}
\label{fig:usvm_example}
\vskip -0.15in
\end{figure}

Learning from contradictions or Universum learning was introduced in \cite{vapnik06} for binary classification problems to incorporate a priori knowledge about admissible data samples. For example, if the goal of learning is to discriminate between handwritten digits `5' and `8', one can introduce additional knowledge in the form of other handwritten letters `a',`b',`c',`d',$\ldots$ `z'. These examples from the Universum  contain certain information about the handwritten styles of authors, but they cannot be assigned to any of the two classes (5 or 8). Further, these Universum samples do not have the same distribution as labeled training samples. In this work we introduce the notion of `Learning from Contradictions' for one class problems. Similar to inductive setting (Definition \ref{def_inductive}) the goal here is also to minimize the generalization error on future test data containing both normal ($y=+1$) and abnormal ($y=-1$) samples. Here however, during training in addition to the samples from the normal class $(\mathbf{x}_i, y_i = +1)_{i=1}^n$, we are also provided with universum (contradictory) samples, which are known not to belong to either of the (normal or abnormal) classes  of interest. A practical use-case can be of automated visual inspection based anomaly detection in manufacturing lines. Here the target is to identify the defects in a specific product type (say 'screws' in Fig. \ref{fig:usvm_example}). For this case, the images from other product types in the manufacturing line act as universum samples. Note that, such universum samples belong to the same application domain (i.e. visual inspection data); but do not represent either of the classes \textcolor{darkgreen}{normal screws} or \textcolor{red}{anomalous screws}. This setting is formalized as,

\begin{definition} {\normalfont (Learning from Contradictions a.k.a Universum Setting)}  \label{def_univsetting}
Given i.i.d training samples $\mathcal{T}=(\mathbf{x}_i,y_i = +1)_{i=1}^n \sim \mathcal{D}_{\mathcal{X}|\mathcal{Y} = +1}^n$, with $\mathbf{x} \in \mathcal{X} \subseteq \Re^d$ and $y \in \mathcal{Y} =\{-1,+1\}$ and additional $m$ universum samples $\mathcal{U} = (\mathbf{x}_{i^\prime}^{*})_{i^\prime=1}^m \sim \mathcal{D}_{\mathcal{U}}$ with $\mathbf{x}^{*} \in \mathcal{X}_{U}^* \subseteq \Re^d$, estimate $h^*:\mathcal{X} \rightarrow \mathcal{Y}$ from hypothesis class $\mathcal{H}$ which, in addition to eq. \eqref{eq_inductive}, obtains maximum contradiction on universum samples i.e. maximizes the following probability for $\mathbf{x}^* \in \mathcal{X}_{U}^*$, 
\begin{flalign} \label{eq_contradition}
&\underset{h \in \mathcal{H}}{\text{sup}} \; \mathbb{P}_{\mathcal{D}_{\mathcal{U}}}[h(\mathbf{x}^* )\notin \mathcal{Y}] = \underset{h \in \mathcal{H}}{\text{sup}} \; \mathbb{E}_{\mathcal{D}_{\mathcal{U}}}[\mathbbm{1}_{ \lbrace \bigcap\limits_{y \in \mathcal{Y}} h(\mathbf{x}^*) \neq y \rbrace}]  &&
\end{flalign}
$\mathcal{D}_{\mathcal{U}}$ is the universum distribution, \\ $\mathbb{P}_{\mathcal{D}_{\mathcal{U}}}(\cdot)$ is   probability under universum distribution,\\
$\mathbb{E}_{\mathcal{D}_{\mathcal{U}}}(\cdot)$ is the  expectation under universum distribution, $\mathcal{X}_{U}^{*}$ is the domain of universum data.
\end{definition}\vspace{-0.3cm}
Learning using contradictions under Universum setting has the dual goal of minimizing the generalization error in \eqref{eq_inductive} while maximizing the contradiction on universum samples \eqref{eq_contradition}. The following proposition provides guidelines on how this can be achieved for the one class hinge loss in \eqref{eq_SVM}. 
\begin{proposition} \label{prop_maxcontradiction}
For the one class hinge loss in \eqref{eq_SVM}, maximum contradiction on universum samples $\mathbf{x}^* \in \mathcal{X}_U^*$ can be achieved when, 
\begin{flalign} \label{max_contradiction}
|\mathbf{w}^\top \phi(\mathbf{x}^{*}) - 1| = 0 
\end{flalign}
\end{proposition}
That is, we need the universum samples to lie on the decision boundary. This motivates the following one class loss using contradictions (under Universum settings) where we relax the constraint in \eqref{max_contradiction} by introducing a $\Delta-$ insensitive loss similar to \cite{weston06,dhar2019} and solve,
\begin{flalign} \label{eq_USVM}
& \underset{\mathbf{w}}{\text{min}}\; \frac{1}{2}||\mathbf{w}||_2^2 +  C \; L_T(\mathbf{w},\phi(\{\mathbf{x}_i\}_{i=1}^n))  &&  \nonumber\\
& \quad \quad \quad \quad \quad \quad + C_U \; L_U(\mathbf{w},\phi(\{\mathbf{x}_{i^\prime}^*\}_{i^\prime=1}^m)) && \\
&\text{s.t.} \quad L_T(\mathbf{w},\phi(\{\mathbf{x}\}_{i=1}^n)) = \sum_{i=1}^n [1-\mathbf{w}^\top\phi(\mathbf{x}_i)]_+&& \nonumber \\
&L_U(\mathbf{w},\phi(\{\mathbf{x}_{i^\prime}^*\}_{i^\prime=1}^m)) = \sum_{i^\prime=1}^m [|1-\mathbf{w}^\top\phi(\mathbf{x}_{i^\prime}^*)|-\Delta]_+ && \nonumber
\end{flalign}
Here, $[x]_+ = \text{max}(0,x)$. Further, the interplay between $C, C_U - $ controls the trade-off between explaining the training samples using $L_T$ vs. maximizing the contradiction on Universum samples using $L_U$. For $C_U = 0$ or $\Delta \rightarrow \infty$, \eqref{eq_USVM} transforms to \eqref{eq_SVM}. For deep learning models, we optimize \eqref{eq_USVM} over all the model parameters and refer to it as Deep One Class Classification using Contradictions (DOC$^3$). 

\subsection{Analysis of Generalization Error bound} \label{sect:generalizationError}
Next we provide theoretical justification in support of Universum learning. We argue, learning under universum settings using DOC$^3$ can provide improved generalization error compared to its inductive counterpart DOC (Hinge). For this, we first derive a generic form of the generalization error bound for one class learning using the Rademacher complexity capacity measure in Theorem \ref{th_oneclass}. 
\begin{theorem} \label{th_oneclass} {\normalfont \textbf{(Generalization Error Bound)}}
Let $\mathcal{F}$ be the class of functions from which the decision function $f(\mathbf{x})$ in eq. \eqref{eq_SVM} and \eqref{eq_USVM} are estimated. Let $R_{f,1} = \{ \mathbf{x}: f(\mathbf{x}) \geq 1 \}$ be the induced decision region. Then, with probability $1-\eta$ with $\eta \in [0,1]$, over any independent draw of the random sample $\mathcal{T}=(\mathbf{x}_i,y_i = +1)_{i=1}^n \sim \mathcal{D}_{\mathcal{T}|\mathcal{Y} = +1}^n$, for any $\kappa > 0$ we have, 
\begin{flalign} \label{eq_generror}
&\mathbb{P}_{\mathcal{D}_{\mathcal{T}|\mathcal{Y} = +1}}(\mathbf{x} \notin R_{f,1-\kappa}) \; \leq \;  \frac{1}{\kappa n} \sum_{i=1}^n \xi_i + \frac{2}{\kappa} \hat{\mathcal{R}}_n(\mathcal{F}) && \nonumber \\
& \quad \quad \quad \quad \quad \quad  \quad \quad \quad \quad \quad \quad   \quad \quad   \quad   \;   + 3 \sqrt{\frac{ln \frac{2}{\eta}}{2n}} && 
\end{flalign}
where, $\quad \xi_i = [1-f(\mathbf{x})]_+$ ; \; $R_{f,\theta} = \{\mathbf{x} : f(\mathbf{x}) \geq \theta \} $ \\
$\hat{\mathcal{R}}_n(\mathcal{F}) = \mathbb{E}_{\sigma}[\underset{f \in \mathcal{F}}{\text{sup}} |\frac{2}{n} \sum_{i=1}^n \sigma_i f(\mathbf{x}_i)| \Big| (\mathbf{x}_i)_{i=1}^n] $\\
$\sigma$ = independent uniform $\{ \pm 1 \}-$ valued random variables a.k.a Rademacher variables.
\end{theorem}
The Theorem \ref{th_oneclass} is agnostic of model parameterization and holds for any popularly adopted kernel machine or deep learning architectures.  Similar to the Theorem 7 in \cite{scholkopf2001estimating}, Theorem \ref{th_oneclass} gives a probabilistic guarantee that new points lie in a larger region $R_{f,1-\kappa}$. Here, we rather use the Empirical Rademacher Complexity (ERC) $\hat{\mathcal{R}}_n(\mathcal{F})$ as the capacity measure of the hypothesis class, instead of the covering number. Additionally, our bound does not contain a $\frac{1}{\kappa^2}$ term as in \cite{scholkopf2001estimating}, and only has the scaling factor of $\frac{1}{\kappa}$. As seen from Theorem \ref{th_oneclass} above, it is preferable to use a hypothesis class $\mathcal{F}$ with smaller ERC $\hat{\mathcal{R}}_n(\mathcal{F})$. Next we compare the ERC of the hypothesis class induced by the formulations \eqref{eq_SVM} vs. \eqref{eq_USVM}.

\begin{theorem} \label{th_rademacher} {\normalfont \textbf{(Empirical Rademacher Complexity)}}. For the hypothesis class induced by the formulations,
\begin{itemize}[nosep]
    \item[--] Eq. \eqref{eq_SVM} : $\mathcal{F}_{\text{ind}} = \{ f: \mathbf{x} \rightarrow \mathbf{w}^{\top} \phi(\mathbf{x}) \Big| ||\mathbf{w}||_2^2 \leq \Lambda^2 \}$ 
    \item[--] Eq. \eqref{eq_USVM} : $\mathcal{F}_{\text{univ}} = \{ f: \mathbf{x} \rightarrow \mathbf{w}^{\top} \phi(\mathbf{x}) \Big| ||\mathbf{w}||_2^2 \leq \Lambda^2 ; |\mathbf{w}^{\top} \phi(\mathbf{x^*}) -1| \leq \Delta \; , \; \forall  x^* \in \mathcal{X}_U^*\}$
\end{itemize}
The following holds,
\begin{enumerate}[label=(\alph*),nosep]
    \item $\hat{\mathcal{R}}_n(\mathcal{F}_{\text{ind}}) \geq \hat{\mathcal{R}}_n(\mathcal{F}_{\text{univ}})$
    \item Further, for any fixed mapping $\phi(\cdot)$,$\; \forall \gamma \geq 0$ we have,
    \begin{enumerate}[leftmargin=0.2cm, label=(\roman*)]
        \item $\hat{\mathcal{R}}_n(\mathcal{F}_{\text{ind}}) \leq \frac{2\Lambda}{n} \sqrt{\sum\limits_{i=1}^n || \mathbf{z}_i||^2}$;\; where $\mathbf{z} = \phi(\mathbf{x})$
        \item $\hat{\mathcal{R}}_n(\mathcal{F}_{\text{univ}}) \leq \frac{2\Lambda}{n} \sqrt{\sum\limits_{i=1}^n || \mathbf{z}_i||^2} \; \underset{\gamma \geq 0} {\text{min}} \;  K(\gamma) \big[1 - \boldsymbol \Sigma(\gamma)\big]  ^{\frac{1}{2}}$ 
    \end{enumerate}
\end{enumerate}
\begin{flalign}
\text{where, }&K(\gamma) = \big[1+\frac{2\gamma m (\Delta^2+1)}{\Lambda^2}\big]^{\frac{1}{2}} &&\\
&\boldsymbol \Sigma(\gamma) = \gamma \frac{ tr(VZ^{\top}ZV^{\top})}{ \big[tr(Z^{\top}Z) \big] \; \big[tr(I+\gamma VV^{\top}) \big]} \label{eq_sig_gamma} && \\
& Z = \begin{bmatrix}
           (\mathbf{z}_1)^T\\
   		   \vdots \\
   		   (\mathbf{z}_{n})^T 
\end{bmatrix} \; and \; V = \begin{bmatrix}
           1\\
           -1
\end{bmatrix} \otimes \begin{bmatrix}
           (\mathbf{u}_1)^T\\
   		   \vdots \\
   		   (\mathbf{u}_{m})^T
\end{bmatrix} && \nonumber \\
& \mathbf{u} = \phi(\mathbf{x}^*); \quad \mathbf{x}^* \in \mathcal{X}_{U}^{*} && \nonumber
% & tr = \text{Matrix Trace}  && \nonumber
\end{flalign}
$\otimes = \text{Kronecker Product}, \quad  tr = \text{Matrix Trace} $ 
\end{theorem}

Note that, several recent works \cite{neyshabur2015norm,sokolic2016lessons,cortes2017adanet} derive the ERC of the function class induced by an underlying neural architecture. In this analysis however, we fix the feature map and analyze how the loss function in \eqref{eq_USVM} reduces the function class capacity compared to \eqref{eq_SVM}. This simplifies our analysis and focuses on the effect of the proposed new loss in \eqref{eq_USVM} under the universum setting. As seen from Theorem \ref{th_rademacher} (a), the function class induced under the universum setting (using contradictions) exhibits lower ERC compared to that under inductive settings. A more explicit characterization of the ERC is provided in part (b). Setting $\gamma = 0$ in (ii), we achieve the same R.H.S as (i); hence the R.H.S in (ii) is always smaller than in (i). Further note that $\boldsymbol \Sigma(\gamma)$ in \eqref{eq_sig_gamma} has the form of a correlation matrix between the training and universum samples in the feature space. In fact, we have $\Sigma(\infty) = \underset{\gamma \rightarrow \infty}{\text{lim}} \Sigma(\gamma) = \frac{ tr(VZ^{\top}ZV^{\top})}{ tr(Z^{\top}Z) \; tr(VV^{\top})}$. This shows that, for a fixed number of universum samples $m$ and $\Delta$, the effect of the DOC$^3$ algorithm is influenced by the correlation between training and universum samples in the feature space. Loosely speaking, the DOC$^3$ algorithm searches for a solution where in addition to reducing the margin errors $\xi_i$, also minimizes this correlation; and by doing so minimizes the generalization error. Similar conclusions have been empirically derived for binary, multiclass problems in \cite{weston06,chapelle2008analysis,cherkassky2011practical,dhar2019}. Here, we provide the theoretical reasoning for one class problems. Further, we confirm these theoretical findings in our results (Section \ref{sec_results_MVTEC_corr}). 

\subsection{Algorithm Implementation} \label{sect:implementation}
A limitation in solving \eqref{eq_USVM} is handling the absolute term in $L_U$. In this paper we adopt a similar approach used in \cite{weston06,dhar2019} and simplify this by re-writing $L_U$ as a sum of two hinge functions. To do this, for every universum sample $\mathbf{x}_{i^\prime}^{*}$  we create two artificial samples, $(\mathbf{x}_{i^\prime}^{*},y_{i^\prime 1}^{*}=1), (\mathbf{x}_{i^\prime}^{*},y_{i^\prime 2}^{*}=-1)$ and re-write, 
\begin{flalign} \label{eq_transform}
& L_U = \sum_{i^\prime=1}^m [|1-\mathbf{w}^\top\phi(\mathbf{x}_{i^\prime}^*)|-\Delta]_+   &&  \\
& = \sum_{i^\prime=1}^m \Big( [\epsilon_1-y_{i^\prime 1}^{*}\mathbf{w}^\top\phi(\mathbf{x}_{i^\prime}^*)]_+  +  [\epsilon_2-y_{i^\prime 2}^{*}\mathbf{w}^\top\phi(\mathbf{x}_{i^\prime}^*)]_+ \Big)   &&  \nonumber
\end{flalign}
where, $\epsilon_1 = 1 - \Delta$ and $\epsilon_2 = -1 - \Delta$. Now, the universum loss is the sum of two hinge functions with $\epsilon_1, \epsilon_2 -$ margins; and can be solved using standard deep learning libraries \cite{pytorch, abadi2016tensorflow,pedregosa2011scikit}. 
\section{Existing Approaches and Related Works}\label{sect:related_works}
Most research in Anomaly Detection (AD) can be broadly categorized as adopting either traditional (shallow) or the more modern deep learning based approaches. Traditional approaches generally adopt parametric or non-parametric statistical modeling, spectral based, or classification based modeling \cite{chandola2009anomaly}. Typical examples include, PCA based methods \cite{jollife07,hoffmann2007kernel}, proximity based methods \cite{knorr2000distance, ramaswamy2000efficient}, tree-based methods like Isolation Forest (IF) \cite{liu2008isolation}, or classification based OC-SVM \cite{scholkopf2001estimating}, Support Vector Data Description (SVDD) \cite{tax2004support} etc. These techniques provide good performance for optimally tuned feature map. However, for complex domains like vision or speech, where designing optimal feature maps is non trivial; such approaches perform sub-optimally. A detailed survey on these approaches is available in \cite{chandola2009anomaly}.

In contrast, for the modern deep learning based approaches, extracting the optimal feature map is imbibed in the learning process. Broadly there are three main sub-categories for deep learning based AD. First, the Deep Auto Encoder and its variants like DCAE \cite{masci2011stacked, makhzani2014winner} or ITAE \cite{huang2019inverse} etc. Here, the aim is to build an embedding where the normal samples are correctly reconstructed while the anomalous samples exhibit high reconstruction error. The second type of approach adopt Generative Adversarial Network (GAN) - based techniques like AnoGAN \cite{schlegl2017unsupervised}, GANomaly \cite{akcay2018ganomaly}, EGBAD \cite{zenati2018efficient}, CBiGAN \cite{carrara2020combining} etc. These approaches, typically focus on generating additional samples which follow similar distribution as the training data. This is followed up by designing an anomaly score to discriminate between normal vs. anomalous samples. Finally, the third category consist of the more recent one class classification based approaches like, DOCC \cite{ruff2018deep}, DROCC \cite{goyal2020drocc} etc. These approaches adopt solving a one class loss function catered for deep architectures. All these above approaches however adopt an unsupervised inductive learning setting. There is a newer class of classification based paradigm which adopts semi or self supervised formulations. Typical examples include, GOAD \cite{bergman2020classification}, SSAD \cite{ruff2019deep}, ESAD \cite{huang2020esad} etc. However, such approaches use fundamentally different problem settings (like a multi class problem for GOAD); or have different assumptions on the additional data available.

% Our DOC$^3$ may seem close to these approaches. However, such approaches use fundamentally different problem settings (like a multi class problem for GOAD); or have different assumptions on the additional data available (i.e. similar data distribution for SSAD or ESAD); and are completely different from the notion of learning from contradictions.
\noindent \textbf{Learning with disjoint auxiliary (DA) data}
A recently popularized new learning setting assumes the availability of an additional auxiliary data which is disjoint from the test set. The underlying assumption is that these auxiliary samples may or may not follow the same distribution as the test data and are disjoint from test set. This idea was first introduced in \cite{dhar2014analysis} (see Section 4.3) and misconstrued as Universum learning. Note that, the notion of universum samples was originally introduced to act as contradictions to the concept classes in the test set \cite{vapnik06}. The above assumption does not adhere to this notion and violates the true essence of Universum learning. This setting has been recently used to propose `outlier exposure' in \cite{hendrycks2018deep} and variants \cite{ruff2021unifying}, \cite{goyal2020drocc}. Our learning from contradiction setting is different from the above methods in the following aspects,
\begin{itemize}[nosep,leftmargin=*]
    \item \textbf{(Problem Setting)} is different. While the above setting only assumes disjoint auxiliary data from test set, Universum follows a different assumption that the concept classes of the universum data is different from both the normal as well as anomalous samples. This assumption is quintessential for proving Prop. \ref{prop_maxcontradiction}, which in turn provides the optimality constraint on the decision function (in eq. \eqref{max_contradiction}). Prop. 2 is not possible for DA setting.
    \item \textbf{(Formulation)} The difference in problem setting is also clear from the  formulations. For example, the formulations proposed under the disjoint auxiliary setting like, \cite{dhar2014analysis}, Outlier Exposure (OE) \cite{hendrycks2018deep} or DROCC-LF \cite{goyal2020drocc} only uses the relation between in-lier training data and the additional auxiliary data. No information on the relation between the auxiliary data and the anomalous samples in test set is encoded in the loss function. In essence, such approaches controls the complexity of hypotheses class by constraining the space in which `normal' samples can lie. In contrast, Universum learning assumes different concept classes for Universum vs. both normal and anomalous (test) samples. This information is encoded through the proof in Prop. \ref{prop_maxcontradiction}. The Universum setting controls the complexity of hypotheses class by constraining the space in which both `normal' or `anomalous' samples can lie. 
\end{itemize}
In short, Universum learning adopts a different learning paradigm (see Definition \ref{def_univsetting}) compared to the `disjoint auxiliary data' settings. Different from the existing `disjoint auxiliary' based loss functions in \cite{dhar2014analysis}, OE \cite{hendrycks2018deep}, DROCC-LF \cite{goyal2020drocc} etc., the Universum samples (in \eqref{eq_USVM}) implicitly contradicts the unseen anomalous test samples. A pedagogical explanation of the differences between these settings with examples is provided in Appendix \ref{sec:OEcomparison}.

\section{Empirical Results} \label{sect:results}
\subsection{Standard Benchmark on CIFAR-10}
\subsubsection{Data Set and Experiment Setup}
For our first set of experiments we use the standard benchmark using the CIFAR-10 \cite{ruff2018deep,goyal2020drocc}. The data consists of 32x32 colour images of 10 classes with 6000 images per class. The classes are mutually exclusive. The underlying task involves one-vs-rest anomaly detection, where we build a one class classifier\begin{wrapfigure}{r}{0.2\textwidth}
\centering
\includegraphics[width=0.18\textwidth]{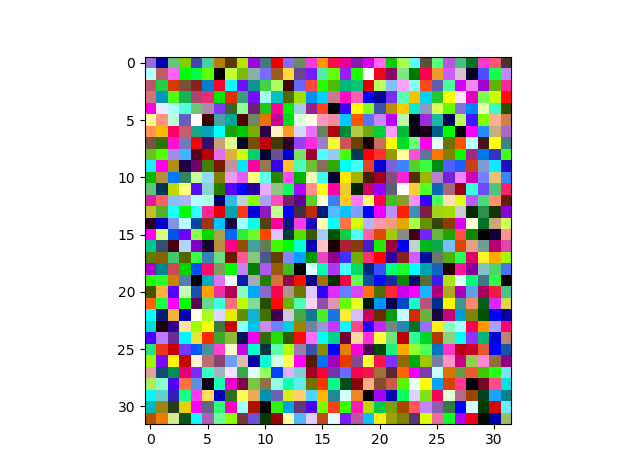}
\vskip -0.1in
\caption{Random noise Universum (contradictions)}.
\label{fig_noiseU}
\vskip -0.15in
\end{wrapfigure} for each class and evaluate  it on the test data for all  the 10-classes. Note that, this data does not have any naturally occurring universum (contradiction) samples (following Def. \ref{def_univsetting}). So, we use synthetic universum samples by randomly generating the pixel values as $\sim \mathcal{N}(\mu,\sigma)$, with $\mu = 0, \; \sigma = 1$; where $\mathcal{N}$ is the normal distribution (see Fig. \ref{fig_noiseU}). The idea of generating synthetic universum (contradiction) samples has been previously studied for binary \cite{weston06,cherkassky2011practical,sinz08}, multiclass \cite{zhang17,dhar2019} and regression \cite{dhar2017} problems. In this paper we use such a similar mechanism for one class problems. Note that for the one-vs-rest AD problem, the generated universum samples do not belong to either `+1' (normal)  or `-1' (anomalous) class used during testing (see Def. \ref{def_univsetting}).The data is scaled in range $[-1,+1]$.  

For this set of experiments we adopt a LeNet like architecture used in \cite{ruff2018deep,goyal2020drocc}. The detailed architecture specifics is provided in Appendix \ref{app_lenet_cifar}.  Note that, this paper focuses on the design and analysis of the DOC$^3$ loss (\eqref{eq_USVM}). Here rather than adopting a state-of-the-art network architecture optimized for the specific dataset; we adopt a systematic approach to isolate the effectiveness of the proposed loss by using a basic LeNet architecture similar to \cite{ruff2018deep, goyal2020drocc}. This avoids secondary generalization effects encoded in most advanced architectures. To that end, the approaches in \cite{ruff2018deep, goyal2020drocc} and their OE extensions serve as the main baselines. We run the experiments over 10 runs.

\subsubsection{Results} \label{sec_results_CIFAR}
\begin{table*}[ht]
\caption{Average AUC (with standard deviation) for one-vs-rest anomaly detection on CIFAR-10. $^\dagger$Results reported in \cite{ruff2018deep}.$^*$Result reported in \cite{goyal2020drocc}. $^\ddagger$Our re-run of the algorithm in \cite{goyal2020drocc}.} \label{table:results1}
\vskip 0.05in
\centering
\resizebox{\textwidth}{!}{%
\begin{tabular}{p{.18\textwidth} p{.1\textwidth} p{.08\textwidth} p{.08\textwidth} p{.08\textwidth} p{.08\textwidth} p{.08\textwidth} p{.08\textwidth} p{.08\textwidth} p{.08\textwidth} p{.08\textwidth}}
% \noalign{
% \hrule height 2pt
% }
\toprule
\textbf{CIFAR-10} & Airplane & Automobile & Bird & Cat & Deer & Dog & Frog & Horse & Ship & Truck \\ %[0.5ex] 
% \noalign{
% \hrule height 2pt
% }
\midrule
\textbf{OC-SVM}$^\dagger$ & 61.6$\pm$0.9 &63.8$\pm$0.6& 50.0$\pm$0.5& 55.9$\pm$1.3 & 66.0$\pm$0.7 & 62.4$\pm$0.8 & 74.7$\pm$0.3 & 62.6$\pm$0.6 & 74.9$\pm$0.4 &  75.9$\pm$0.3 \\ 
\textbf{IF}$^\dagger$ & $60.1 \pm 0.7$& $50.8 \pm 0.6$ & 49.2$\pm$0.4 & 55.1$\pm$0.4 & 49.8$\pm$0.4 & 58.5$\pm$0.4 &42.9$\pm$0.6 & 55.1$\pm$0.7& 74.2$\pm$0.6 & 58.9$\pm$0.7 \\
\textbf{DCAE}$^\dagger$ & $59.1 \pm 5.1$& $57.4 \pm 2.9$ & 48.9$\pm$2.4 & 58.4$\pm$1.2 & 54.0$\pm$1.3 &62.2$\pm$1.8 &51.2$\pm$5.2 & 58.6$\pm$2.9& 76.8$\pm$4.1 & 67.3$\pm$3.0 \\
\textbf{AnoGAN}$^\dagger$ & $67.1 \pm 2.5$& $54.7 \pm 3.4$ & 52.9$\pm$3.0 & 54.5$\pm$1.9 & 65.1$\pm$3.2 & 60.3$\pm$2.6 &58.5$\pm$1.4 & 62.5$\pm$0.8& 75.8$\pm$4.1 & 66.5$\pm$2.8 \\
\textbf{ConAD 16}$^*$ & $77.2$& $63.1$ & 63.1& 61.5 & 63.3 & 58.8 & 69.1 & 64 & 75.5 & 63.7 \\
\textbf{1-NN}$^*$ & $69.02$& $44.2$ & 68.27 & 51.32 & \textbf{76.71} & 49.97 &72.44 & 51.13& 69.09 & 43.33 \\
\textbf{DOCC} (Soft-Bound)$^\dagger$ & $61.7 \pm 4.2$& $64.8 \pm 1.4$ & 49.5$\pm$1.4 & 56.0$\pm$1.1 & 59.1$\pm$1.1 & 62.1$\pm$2.4 &67.8$\pm$2.4 & 65.2$\pm$1.0& 75.6$\pm$1.7 & 71.0$\pm$1.1 \\
\textbf{DOCC}$^\dagger$ & $61.7 \pm 4.1$& $65.9 \pm 2.1$ & 50.8$\pm$0.8 & 59.1$\pm$1.4 & 60.9$\pm$1.1 & 65.7$\pm$2.5 &67.7$\pm$2.6 & 67.3$\pm$0.9& 75.9$\pm$1.2 & 73.1$\pm$1.2 \\
\textbf{DROCC}$^\ddagger$ & $79.2\pm1.9$& $\mathbf{74.9\pm2.6}$ & 68.3$\pm$1.5 & $\mathbf{62.3\pm2.7}$ & 70.3$\pm$2.7 & \textbf{66.1}$\pm$\textbf{2.0} &68.1$\pm$2.2 & \textbf{71.3}$\pm$\textbf{4.6}& 62.3$\pm$10.3 & \textbf{76.6}$\pm$\textbf{1.9} \\
\textbf{DOC} (Hinge eq. \eqref{eq_SVM}) & $76.8 \pm 1.4$& $62.6 \pm 2.8$ & 52.1$\pm$0.7 & 60.4$\pm$0.7 & 62.3$\pm$1.2 & 61.9$\pm$1.9 &76.3$\pm$0.5 & 59.8$\pm$1.3& 72.8$\pm$1.1 & 74.9$\pm$2.0 \\
\textbf{DOC} (DA/OE) & $69.5 \pm 14.5$ & $73.1 \pm 3.3$ & $67.3\pm0.3$ & $\mathbf{62.4\pm2.5}$ & 71.1$\pm$2.1 & $\mathbf{67.3 \pm 7.4}$ & $\mathbf{78.6\pm1.7}$ & 66.8$\pm$2.4& $70.3\pm1.9$ & $75.8\pm2.5$\\
\textbf{DOC}$^3$ (univ) & $\mathbf{81.3 \pm 0.5}$ & $\mathbf{74.2 \pm 1.3}$ & $\mathbf{69.0\pm0.6}$ & $\mathbf{62.1\pm0.4}$ & 74.0$\pm$1.6 & 63.0$\pm$4.6 &$\mathbf{77.7\pm0.3}$ & 67.6$\pm$1.8& $\mathbf{81.1\pm0.6}$ & $\mathbf{76.8\pm2.0}$\\ \midrule
\textbf{DROCC-LF}(OE) & $91.9 \pm 0.9$ & $70.5 \pm 2.4$ & $70.9\pm2.2$ & $63.1\pm2.2$ & 76.6$\pm$1.3 & 65.7$\pm$1.4 & $74.1\pm2.7$ & 70.6$\pm$3.6& $85.1\pm4.1$& $84.6\pm2.4$\\
\textbf{DROCC-LF}(univ) & $96.8 \pm 0.4$ & $88.2 \pm 5.3$ & $79.8 \pm 1.6$  & $62.7 \pm 5.3$ & $80.4 \pm 1.3$  & $84.9 \pm 4.1$ &$87.4 \pm 0.9$  &  $75.0 \pm 1.4$ & $93.4 \pm 0.2$ & $86.5 \pm 0.8$\\
% \noalign{
% \hrule height 2pt
% }
\bottomrule
\end{tabular}}
\vskip -0.15in
\end{table*}

Table \ref{table:results1} provides the average $\pm$ standard deviation of the AUC under the ROC curve over 10 runs of the experiment. Here, we report the results of the best performing DOC (Hinge in \eqref{eq_SVM}) model selected over the range of parameters $\lambda = 1/2C  = [1.0, 0.5]$ and that for DOC$^3$ over the range of parameters $\lambda = 1/2C = [0.1, 0.05], C_{U}/C = [1.0, 0.5]$. Through out the paper we fix $\Delta = 0$. A more detailed discussion on model selection and the selected model parameters is provided in Appendix \ref{app_mod_sel_cifar}. Moreover, for a more thorough comparison we also include the results of the DOC using Hinge loss extended under the Disjoint Auxiliary (DA) or Outlier Exposure (OE) setting. Throughout the paper, as an exemplar for the DA/OE setting; we use the additional universum  samples as belonging to the negative class following  \cite{goyal2020drocc}. In addition we also report the benchmark results for both shallow and deep learning methods provided in \cite{ruff2018deep,goyal2020drocc}. Note however, our results for the DROCC algorithm is different from that reported in \cite{goyal2020drocc}. Re-running the codes provided in \cite{goyal2020drocc} did not yield similar results as reported in the paper (especially for `Ship'). Moreover, their current implementation normalizes the data using mean, $\mu = (0.4914,0.4822, 0.4465)$ and standard deviation, $\sigma= (0.247, 0.243, 0.261)$. These values are calculated using the data from all the classes; which is not available during training of a single class. To avoid such inconsistencies we rather normalize using mean, $ \mu = (0.5, 0.5, 0.5)$ and standard deviation, $\sigma=(0.5, 0.5, 0.5)$. Such a scale does not need apriori information of the other class's pixel values and scales the data in a range of $[-1,+1]$. Detailed discussions on reproducing the results of the recent deep learning algorithms  DOCC \cite{ruff2018deep} and DROCC \cite{goyal2020drocc} is provided in Appendix \ref{app_docc_drocc_reproduce}.

As seen from Table \ref{table:results1}, DOC$^3$ (using the noise universum), provides significant improvement $\sim 5 - 15 \%$ (and upto $30 \%$ for `Bird'), over its inductive counterpart (DOC). In addition, the DOC$^3$ in most cases outperforms the DOC (DA/OE). This illustrates the advantage of extending Anomaly Detection problems following Def. \ref{def_univsetting} in accordance with the Prop. \ref{prop_maxcontradiction}. To further consolidate our approach we compare the advanced adversarial based DROCC-LF method \cite{goyal2020drocc} (under OE settings) vs. our extension of DROCC-LF under universum setting. The major difference is now the auxilliary data serves as universum samples and the loss function follows \eqref{eq_USVM} (see Appendix \ref{app_drocc_lf} Algo. \ref{alg:drocclf-nn}). As seen from Table \ref{table:results1} the DROCC-LF (univ) significantly outperforms the DROCC-LF (OE) method to upto - 30\% (`dog') for some cases. Details on the experiment setup and the optimal model parameters are available in Appendix \ref{app_drocc_lf} for reproducibilty. Additional results showing DOC$^3$ improving the state-of-the-art for tabular data sets Abalone, Arrhythmia, Thyroid used in \cite{goyal2020drocc} is provided in Appendix \ref{app_tab_results}.

%Of course, the performance of DOC$^3$ depends on several factors like, selection of optimal network architecture, model parameters and `good' universum data.   

%\subsection{Visual Inspection based Anomaly Detection}
\subsection{Visual Inspection using MV-Tec AD data}
For our next set of experiments we tackle the more realistic visual inspection based anomaly detection problem in manufacturing lines. Lately with the recent advancements in deep learning technologies, there has been an increased interest towards automating manufacturing lines and adopting AI driven solutions providing automated visual inspection of product defects \cite{bergmann2019mvtec, huang2015automated}. One popular benchmark data set used for such problems is the MV-Tec AD data set \cite{bergmann2019mvtec}.    
\subsubsection{Data Set and Experiment Setup} \label{sec_mvtec_exp}
The MV-Tec AD data set contains 5354 high-resolution color images of different industrial object and texture categories. For each categories it contains normal (no defect) images used for training. The test data contains both normal as well as anomalous (defective) product images. The anomalies manifest themselves in the
form of over 70 different types of defects such as scratches,
dents, contamination, and various other structural changes. The goal in this paper is to build one class image-level classifiers for the texture categories (see Table \ref{tab_MVtecData}). We use the original data scale of [0,1]. Further, to simplify the problem we resize all the images to $64 \times 64$ pixel. Note that, for the current analysis we only use the texture classes containing RGB images.  

For this problem we have naturally occurring universum (contradiction) samples in the form of the objects' images or other texture types. That is, for the goal of building a one class classifier for `carpet', all the `\textit{other textures}' (leather, tile, wood) or the `\textit{objects}' (bottle, cable, capsule, hazelnut, metal nut, pill, transistor) available in the dataset, can serve as universum (contradiction) samples. This is inline with the problem setting in Def. \ref{def_univsetting}, where such samples are neither `normal' nor `anomalous' (defective) carpet samples. For our experiments, we use three types of universum,
\begin{itemize}[nosep,leftmargin=*]
    \item \textbf{Noise}: Similar to previous experiments we generate random noise as universum samples. Here, since the data is already scaled in the range of [0,1], we generate $64 \times 64$ dimension images where the pixel values are obtained from a uniform distribution $\sim \mathcal{U}(0,1)$.
    \item \textbf{Objects}: This type of universum contains all the images in the object categories with RGB pixels viz. bottle, cable, capsule, hazelnut, metal nut, pill, transistor. Note that, we include both the normal as well as the defective samples for these objects.
    \item \textbf{other Textures}: Here we use the remaining texture images as universum. That is, if the goal is building a one class classifier for `carpet' we use the images from the other `textures' (leather, tile, wood) as universum. We include both the normal as well as the defective samples in the universum set.
\end{itemize}
As before, we adopt a LeNet like architecture (schematic representation in Fig. \ref{fig_net_arc_mvtec2}, details in Appendix \ref{app_lenet_mvtec}).  Note that, there have been a few recent works proposing advanced architectures to achieve state-of-the-art performance on this data \cite{carrara2020combining,huang2019attribute}. However, the main focus here is to isolate the effectiveness of DOC3, and hence we mainly compare against DOC and DOC(OE) baselines using a simple LeNet network. Since our baselines DOC, DOC(OE) using LeNet have not been previously reported on this data; as sanity check we also add the results in \cite{massoli2020mocca} for a good comparison with different classes of algorithms. Also, we adopt a slight modification to our loss function. Rather than using relu function $[x]_+$ in \eqref{eq_SVM}, and \eqref{eq_USVM} for the training samples; we  use a softplus operator. We see improved results using this modification. Note that, softplus is a dominating surrogate loss over relu, and hence Theorem \ref{th_oneclass} still holds.   

\begin{table}[t]
\caption{MVTec-AD Dataset} \label{tab_MVtecData}
\vskip 0.08in
\begin{center}
\begin{small}
\begin{sc}
\begin{tabular}{lccc}
\toprule
\multirow{2}{*}{Textures} & \multirow{2}{*}{Train }& \multicolumn{2}{c}{Test} \\
& &Normal& Anomaly\\
\midrule
Carpet    & 280 & 28 & 89 \\
Leather & 245& 32& 92\\
Tile    & 230& 33 & 84 \\
Wood    & 247 & 19 &  60  \\
\bottomrule
\end{tabular}
\end{sc}
\end{small}
\end{center}
\vskip -0.25in
\end{table}

\subsubsection{Performance comparison results} \label{sec_results_MVTEC}

\begin{table*}[ht]
\caption{AUC for MVTec-AD (Texture) data. $^\dagger$ Results taken from \cite{massoli2020mocca}. \textbf{Bold} = best overall model. \underline{Underline} = best universum or OE model.} \label{tab_MVtecAUC}
%\vskip -0.7in
\begin{center}
\resizebox{\textwidth}{!}{
% \begin{small}
% \begin{sc}
% \tabcolsep = 0.1cm
\begin{tabular}{p{0.1\textwidth}|p{0.02\textwidth}p{0.02\textwidth}p{0.02\textwidth}p{0.02\textwidth}p{0.02\textwidth}p{0.02\textwidth}|p{0.08\textwidth}|p{0.08\textwidth}|p{0.08\textwidth}|p{0.08\textwidth}||p{0.08\textwidth}|p{0.08\textwidth}|p{0.08\textwidth}}
\toprule
\multirow{3}{*}{\sc{Textures}} & \multirow{3}{*}{\rotatebox{90}{\small{AE$_{L2}^\dagger$} }}& \multirow{3}{*}{ \rotatebox{90}{\small{GeoTrans$^\dagger$}}}& \multirow{3}{*}{\rotatebox{90}{\scriptsize{GANomaly$^\dagger$}}} & \multirow{3}{*}{\rotatebox{90}{\small{ITAE$^\dagger$}}} & \multirow{3}{*}{\rotatebox{90}{\small{EGBAD$^\dagger$}}} & \multirow{3}{*}{\rotatebox{90}{\small{CBiGAN$^\dagger$}}} & \specialcell{DOC \eqref{eq_SVM}} &
\specialcell{DOC (OE) \\ (noise)} &
\specialcell{DOC (OE) \\ (objects)} &
\specialcell{DOC (OE) \\ (textures)} &
\specialcell{DOC$^3$ \\ (noise)} &
\specialcell{DOC$^3$ \\ (objects)} &
\specialcell{DOC$^3$ \\ (textures)} 
\\ \cline{8-14}
& & & & & & & \specialcell{Best \\ Avg. $\pm$ std} & \specialcell{Best \\ Avg. $\pm$ std}&\specialcell{Best \\ Avg. $\pm$ std} &\specialcell{Best \\ Avg. $\pm$ std} &\specialcell{Best \\ Avg. $\pm$ std} &\specialcell{Best \\ Avg. $\pm$ std} 
&\specialcell{Best \\ Avg. $\pm$ std}
\\
\midrule
Carpet & 64 & 44 & 70 & 71 & 52 & 55 & \specialcell{$81.1$\\\small{$81.1\pm0.0$}} 
&\specialcell{$76.2$\\\small{$56.5\pm10.1$}}  &\specialcell{$89.6$\\\small{$82.1\pm4.2$}} &\specialcell{$54.9$\\\small{$49.2\pm4.9$}} &\specialcell{\textbf{\underline{95.7}}\\\small{$80.4\pm 8.4$}} &\specialcell{93.8\\\small{\underline{$\mathbf{87.5\pm 3.8}$}}} & \specialcell{81.1\\\small{$81.1\pm0.0$}} 
\\ \hline
Leather & 80 & 84&84 &86 &55 &83&  \specialcell{$63.1$\\\small{$62.7\pm0.3$}} & \specialcell{$65.7$\\\small{$64.8\pm0.8$} } &  \specialcell{\underline{$\mathbf{95.5}$}\\\underline{\small{$\mathbf{89.6\pm5.1}$}}} &  \specialcell{$40.1$\\\small{$39.8\pm0.2$}} & \specialcell{$88.1$\\\small{$82.9\pm 4.5$}} &\specialcell{$93.5$\\\small{$83.1\pm7.5$}} & \specialcell{$63.1$\\\small{$62.4\pm0.5$}}\\ \hline
Tile & 74 & 42 & 79 & 74 & 79 & \textbf{91}&  \specialcell{$62.8$\\\small{$62.3 \pm 0.7$}} & \specialcell{$65.9$\\\small{$64.9\pm0.7$}}& 
\specialcell{$75.7$\\\small{$74.0\pm1.4$}}&
\specialcell{$67.7$\\\small{$65.3\pm0.9$}}&
\specialcell{$66.3$\\\small{$64.7\pm0.6$}}&
\specialcell{\underline{$77.0$}\\\underline{\small{$76.5\pm0.5$}}} & \specialcell{$65.1$\\\small{$64.4\pm0.5$}} \\ \hline
Wood & \textbf{97}& 61& 83 & 92 & 91 & 95 & \specialcell{$41.1$\\\small{$40.6\pm0.1$}} & \specialcell{$90.2$\\\small{$82.8\pm5.6$}} & 
\specialcell{$77.2$\\\small{$70.9\pm6.7$}}&
\specialcell{$52.6$\\\small{$50.5\pm1.5$}}&
\specialcell{\underline{$93.1$}\\ $\underline{83.4\pm 7.0}$} &\specialcell{$75.3$\\\small{$69.0\pm5.8$}}  &\specialcell{$49.4$\\\small{$49\pm0.4$}} \\ 
\bottomrule
\end{tabular}
% \end{sc}
% \end{small}
}
\end{center}
\vskip -0.2in
\end{table*}

Table \ref{tab_MVtecAUC} provides the results over 10 runs of our experiments. We provide the the average $\pm$ standard deviation of the AUC values for DOC, DOC (DA/OE) and DOC$^3$ algorithm. In addition we also provide the best AUC obtained for each algorithm over these 10 runs. Additional details on model selection and the optimal hyperparameters is provided in Appendix \ref{app_mod_sel_mvtec}. As seen in Table \ref{tab_MVtecAUC}, the DOC$^3$ algorithm provides significant improvement over DOC. Depending on the type of universum typical improvements range upto $> 50 \%$.  In addition, DOC$^3$ provides consistent improvements over the DOC (DA/OE) algorithm. In all, these results further consolidate the utility of DOC$^3$ under the universum setting (Def \ref{def_univsetting}). Separately, Table \ref{tab_MVtecAUC} also provides the baseline results available in \cite{massoli2020mocca}.  Note that, these results are obtained using advanced network architectures adopted for the MVTec data, and are not averaged over multiple runs. Hence, we compare these results with the best AUC obtained for DOC, DOC (DA/OE) and DOC$^3$ over 10 runs. As seen from Table \ref{tab_MVtecAUC}, DOC$^3$ improves upon the `carpet' and `leather' results using the `objects' universum. Further, it achieves comparable performance for the `Wood' and `Tile' texture using `Noise' and `Obj.' universum respectively. Achieving  improved performance over the baseline algorithms, even using a basic LeNet architecture sheds a very positive note for the proposed DOC$^3$ algorithm. 
% For the DOC algorithm we select the best model over the range of parameters $\lambda = 1/2C = [1.0, 0.1, 0.01, 0.005]$, and for DOC$^3$ over the range $\lambda$ (as DOC above) and $C_{U}/C = [2.0, 1.0, 0.1, 0.01]$, $\Delta =0$.

\subsubsection{Understanding DOC$^3$ performance using Theorem \eqref{th_rademacher}} \label{sec_results_MVTEC_corr}

\begin{figure}
\centering
\includegraphics[width=0.5\textwidth]{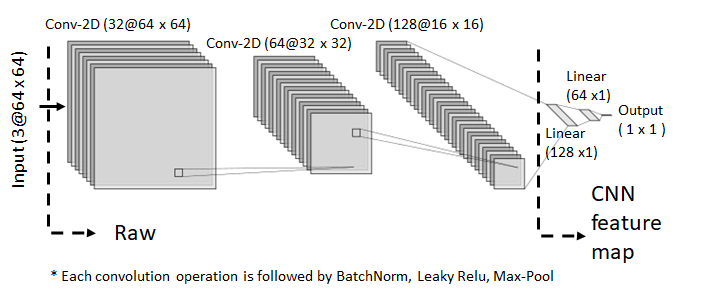}
\vskip -0.1in
\caption{Schematic representation of the Network used for MVTec-AD results in Table \ref{tab_MVtecAUC}.}
\label{fig_net_arc_mvtec2}
\vskip -0.15in
\end{figure}

For our final set of experiments we try to understand the working of the DOC$^3$ algorithm in connection with the correlation $\Sigma(\infty)$ (in Theorem \ref{th_rademacher}). Table \ref{tab_MVtecCorr} reports the correlation values for the training and universum samples using `RAW' pixel, `DOC' and DOC$^3$ solution's feature maps. For the feature map we use the CNN features shown in Fig. \ref{fig_net_arc_mvtec2}. Also, the DOC$^3$ solutions represent the estimated model using the training data (in column 1) and the respective universum data (in column 2). As seen from the results, the DOC solution provides high correlation $\Sigma(\infty)$ between the training and universum samples. In essence, the DOC solution sees the training and universum samples similarly. This is not desirable, as the universum samples follow a different distribution than training samples. On the contrary, the DOC$^3$ provides a solution where the correlation between the training  and universum samples are significantly reduced. This is inline with the Theorem \ref{th_rademacher}'s analysis (section \ref{sect:generalizationError}), where we argued that the DOC$^3$ searches for a solution with low $\Sigma(\infty)$ between the training and universum samples (in feature space). And by doing so ensures lower ERC and improved generalization compared to DOC (confirmed empirically in Table \ref{tab_MVtecAUC}). Another interesting point seen for the `other texture' universum type, with originally high raw pixel correlation values ($\sim 0.9$) is that; using DOC$^3$ provides limited improvement. Such universum types are too similar to the training data, and act as `bad' contradictions.    

\begin{table}[t]
\caption{Average $\pm$ standard deviation of correlation $\Sigma(\infty)$ between training and universum over 10 runs. Values scaled $\times 10^2$} \label{tab_MVtecCorr}
\vskip 0.15in
\begin{center}
\resizebox{0.95\columnwidth}{!}{
\begin{small}
\begin{sc}
% \tabcolsep = 0.1cm
%\resizebox{0.46\textwidth}{!}{
\begin{tabular}{c|c|ccc}
\toprule
\specialcell{Train \\Data} & \specialcell{Univ.\\ Data} & Raw & \specialcell{DOC\\ \small{ Feat. map}} & \specialcell{DOC$^3$ \\ \small{ Feat. map}}\\ \midrule
\multirow{3}{*}{Carpet} & Noise & $73.8$ & $99.9\pm0.0$ &  $17.5\pm3.8$  \\
&Obj. & $73.8$  & $97.1\pm 0.1$ & $28.8\pm3.4$  \\
&Text. & $92.4$ & $99.9\pm0.0$ & $68.7\pm0.0$ \\ \hline
\multirow{3}{*}{Leather} & Noise & $70.1$ & $99.3\pm0.1$ &  $42.1\pm1.9$  \\
&Obj. & $70.7$  & $91.9\pm0.3$ & $23.5\pm4.3$  \\
&Text. & $93.6$ & $99.3\pm0.1$ & $97.0\pm1.7$ \\ \hline
\multirow{3}{*}{Tile} & Noise & 71.2 & $99.8\pm0.0$ & $54.2\pm 1.0$ \\
&Obj. & 71.5 & $92.8\pm0.8$ & $31.3\pm8.8$  \\
&Text. & 89.9  & $99.6\pm0.0$ & $89.3\pm2.2$ \\ \hline
\multirow{3}{*}{Wood} & Noise & $69.5$& $99.8\pm0.0$ & $39.8\pm5.8$  \\
&Obj. &$69.8$ & $93.5\pm0.2$ & $41.9\pm7.1$\\
&Text. &$91.7$ & $99.7\pm0.0$ & $64.4\pm2.1$\\
\bottomrule
\end{tabular}
%}
\end{sc}
\end{small}}
\end{center}
\vskip -0.1in
\end{table}

\section{Conclusions}
This paper introduces the notion of learning from contradictions for deep one class classification and introduces the DOC$^3$ algorithm. DOC$^3$ is shown to provide improved generalization over DOC, its inductive counterpart, by deriving the Empirical Rademacher Complexity (ERC). We empirically show the effectiveness of the proposed formulation, and connect the results to our theoretical analysis. Finally, we also discuss the limitations and the future research directions (moved to Appendix \ref{sec_futureresearch} due to space constraints).

% Optimal model tuning is of utmost importance for large-margin loss using deep architectures because we see similar data piling effects onto the margin borders previously reported as `maximum data piling' (MDP) for traditional approaches \cite{ahn2010maximal,cherkassky2011practical} (see Appendix \ref{app_mdp}). This, typically leads to high variation in generalization performance.

\bibliography{doc3icml2022}
\bibliographystyle{icml2022}

%%%%%%%%%%%%%%%%%%%%%%%%%%%%%%%%%%%%%%%%%%%%%%%%%%%%%%%%%%%%%%%%%%%%%%%%%%%%%%%
%%%%%%%%%%%%%%%%%%%%%%%%%%%%%%%%%%%%%%%%%%%%%%%%%%%%%%%%%%%%%%%%%%%%%%%%%%%%%%%
% APPENDIX
%%%%%%%%%%%%%%%%%%%%%%%%%%%%%%%%%%%%%%%%%%%%%%%%%%%%%%%%%%%%%%%%%%%%%%%%%%%%%%%
%%%%%%%%%%%%%%%%%%%%%%%%%%%%%%%%%%%%%%%%%%%%%%%%%%%%%%%%%%%%%%%%%%%%%%%%%%%%%%%
\clearpage
\appendix
%\appendix
%\section{Table of Notations} \label{sec:ton}
\section{Proofs} \label{sec:proofs}
\subsection{Proof of Proposition \ref{prop_equivalence}} \label{sec:proof_prop_equivalence}
\noindent \underline{Part i}
A slightly different version of this proposition is analyzed in Proposition 8.2 of \cite{scholkopf2002learning} and \cite{chang2001training}. Here, we provide a different version of the connection between the solutions of \eqref{eq_SVM} and \eqref{eq_nuSVM}. This is achieved through analyzing the KKT systems of the formulations. We start with the formulation \eqref{eq_SVM}. Note that, \eqref{eq_SVM} is the same as solving,
\begin{flalign}
&\underset{\mathbf{w}}{\text{min}} \; \frac{1}{2} ||\mathbf{w}||^2 + C \sum_{i=1}^n \xi_i &&\\
&\text{s.t.} \; \mathbf{w}^\top \phi(\mathbf{x}_i) \geq 1 - \xi_i ; \; \xi_i \geq 0&& \nonumber
\end{flalign}
The Lagrangian is given as, \\ 
$\mathcal{L}(\mathbf{w},\xi,\alpha,\beta) = \frac{1}{2} ||\mathbf{w}||^2 + C \sum_{i=1}^n \xi_i - \sum_{i=1}^n \beta_i \xi_i -\sum_{i=1}^n \alpha_i[\mathbf{w}^\top \phi(\mathbf{x}_i) -1 + \xi_i]$ \\ \\
\underline{KKT System} 
\begin{flalign}
& \nabla_{\mathbf{w}} \mathcal{L} = 0 \Rightarrow \mathbf{w} = \sum_{i=1}^n \alpha_i\phi(\mathbf{x}_i) &&  \label{eq_kkt1} \\
& \nabla_{\xi} \mathcal{L} = 0 \Rightarrow C = \alpha_i + \beta_i &&
\end{flalign}
\noindent Complimentary Slackness,
\begin{flalign}
& \alpha_i[\mathbf{w}^\top \phi(\mathbf{x}_i) - 1+\xi_i] = 0&&\\
& \beta_i \xi_i = 0&& 
\end{flalign}
\noindent Constraints,
\begin{flalign}
& \mathbf{w}^\top \phi(\mathbf{x}_i) \geq 1-\xi_i &&\\
& \xi_i \geq 0 \label{eq_kkt1end} && 
\end{flalign}

\noindent Define $ \delta = \frac{1}{\sum_i \alpha_i}$ and re-write the equations \eqref{eq_kkt1} - \eqref{eq_kkt1end} by scaling with $\delta > 0$ \; as, $\mathbf{\hat w} = \mathbf{w}\delta; \hat\alpha_i = \alpha_i\delta; \; \hat\beta_i = \beta_i\delta; \; \hat\xi_i = \xi_i \delta$. ($\delta > 0; \because \exists i\; \text{s.t.} \; \alpha_i >0 \text{ and } \forall i, \alpha_i \geq 0)$ \\
This gives, \underline{Transformed KKT System} 
\begin{flalign}
& \mathbf{\hat w} = \sum_{i=1}^n \hat \alpha_i\phi(\mathbf{x}_i) &&  \label{eq_kkt2} \\
& C\delta = \hat\alpha_i + \hat\beta_i &&
\end{flalign}
\noindent Complimentary Slackness,
\begin{flalign}
& \hat \alpha_i[\mathbf{\hat w}^\top \phi(\mathbf{x}_i) - \delta+\hat \xi_i] = 0&&\\
& \hat \beta_i \hat \xi_i = 0 && 
\end{flalign}
\noindent Constraints,
\begin{flalign}
& \mathbf{\hat w}^\top \phi(\mathbf{x}_i) \geq \delta-\hat \xi_i ; \quad \hat \xi_i \geq 0 \label{eq_kkt2end} &&
\end{flalign}
Note that, the transformed KKT system \eqref{eq_kkt2} - \eqref{eq_kkt2end} solves \eqref{eq_nuSVM} with $\nu = \frac{1}{Cn\delta}; \rho = \delta$  (compare with the  KKT of \eqref{eq_nuSVM}).

\noindent \underline{Part ii}
For the solution to \eqref{eq_nuSVM} obtained from Proposition \ref{prop_equivalence} (i) the decision rule can be given as,
\begin{flalign}
&\mathbf{\hat w}^\top \phi(\mathbf{x}) - \rho = 0 \Rightarrow (\mathbf{w}\delta)^\top \phi(\mathbf{x}) - \delta = 0 && \nonumber \\ 
&\Rightarrow \mathbf{w}^\top \phi(\mathbf{x}) - 1 = 0 \quad (\because \delta > 0) \nonumber &
\end{flalign} \qed

\subsection{Proof of Proposition \ref{prop_maxcontradiction}} \label{sec:proof_prop_maxcontradiction}
\noindent Note that for this proof we need to accommodate a case where a sample may not belong to either of the two classes $\{-1, +1\}$. For this we rather analyze a different decision rule than \eqref{eq_SVM}. \\
\noindent Define,
\[g(\mathbf{x}) = \left\{
\begin{array}{l l}
    +1;\quad  \text{if} \; \mathbf{w}^\top \phi(\mathbf{x}_i) > 1  \\
    -1;\quad  \text{if} \; \mathbf{w}^\top \phi(\mathbf{x}_i) < 1  \\
\end{array}\right. \]

\noindent This gives, 
\begin{flalign}
&g(\mathbf{x}) = +1 \Rightarrow h(\mathbf{x}) = +1  \; \text{and} \;  g(\mathbf{x}) = -1 \Rightarrow h(\mathbf{x}) = -1 && \nonumber \\
&\Rightarrow \mathbbm{P}_{\mathcal{D}_\mathcal{U}}(h(\mathbf{x}^*) = y) \geq \mathbbm{P}_{\mathcal{D}_\mathcal{U}}(g(\mathbf{x}^*) = y) \; \forall y = \{-1,+1\} && \nonumber
\end{flalign}
Since the events are mutually exclusive we have,
\begin{flalign}
&\mathbb{E}_{\mathcal{D}_{\mathcal{U}}}\left(\bigcup\limits_{y \in \{-1,+1\}}  \mathbbm{1}[h(\mathbf{x}^*) = y] \right) && \nonumber \\ 
& \quad \quad \quad \quad  \quad \quad \quad \quad  \geq \mathbb{E}_{\mathcal{D}_{\mathcal{U}}} \left(\bigcup\limits_{y \in \{-1,+1\}}  \mathbbm{1}[g(\mathbf{x}^*) = y] \right) && \nonumber 
\\
&\Rightarrow \underset{h}{\text{inf}}\; \mathbb{E}_{\mathcal{D}_{\mathcal{U}}} \left(\bigcup\limits_{y \in \{-1,+1\}}  \mathbbm{1}[h(\mathbf{x}^*) = y] \right) && \nonumber \\ & \quad \quad \quad \quad  \quad \quad \quad \quad   \geq \underset{g}{\text{inf}} \; \mathbb{E}_{\mathcal{D}_{\mathcal{U}}} \left(\bigcup\limits_{y \in \{-1,+1\}}  \mathbbm{1}[g(\mathbf{x}^*) = y] \right) && \nonumber \\
& \Rightarrow 1-\underset{h}{\text{inf}} \; \mathbb{E}_{\mathcal{D}_{\mathcal{U}}} \left(\bigcup\limits_{y \in \{-1,+1\}}  \mathbbm{1}[h(\mathbf{x}^*) = y] \right) && \nonumber \\ & \quad  \quad \quad \quad \quad \quad \quad \quad  \leq 1-\underset{g}{\text{inf}} \; \mathbb{E}_{\mathcal{D}_{\mathcal{U}}} \left(\bigcup\limits_{y \in \{-1,+1\}}  \mathbbm{1}[g(\mathbf{x}^*) = y] \right) && \nonumber \end{flalign}
\begin{flalign}
&\Rightarrow \underset{h}{\text{sup}} \; \mathbb{E}_{\mathcal{D}_{\mathcal{U}}} \left(\bigcap\limits_{y \in \{-1,+1\}}  \mathbbm{1}[h(\mathbf{x}^*) \neq y] \right) \; \text{\small (De-Morgan's law)} && \nonumber \\ & \quad \quad \quad \quad   \leq \underset{g}{\text{sup}} \; \mathbb{E}_{\mathcal{D}_{\mathcal{U}}} \left(\bigcap\limits_{y \in \{-1,+1\}}  \mathbbm{1}[g(\mathbf{x}^*) \neq y] \right) && \nonumber \\
& \quad \quad \quad \quad  = \underset{g}{\text{sup}} \; \mathbb{E}_{\mathcal{D}_{\mathcal{U}}} \left( \mathbbm{1}[g(\mathbf{x}^*) \neq +1] \land \mathbbm{1}[g(\mathbf{x}^*) \neq -1] \right) && \nonumber \\
& \quad  = \underset{g}{\text{sup}} \; \mathbb{E}_{\mathcal{D}_{\mathcal{U}}} \left( \mathbbm{1}[\mathbf{w}^\top \phi(\mathbf{x})  -1 \ngtr 0] \land \mathbbm{1}[\mathbf{w}^\top \phi(\mathbf{x}) -1 \nless 0]  \right) && \nonumber
\end{flalign}
The maximum can be achieved when $\mathbf{w}^\top \phi(\mathbf{x}) -1  = 0$ \qed

\subsection{Proof of Theorem \ref{th_oneclass}} \label{sec:proof_th_oneclass}
\noindent Define, $R_{f,\theta} = \{\mathbf{x} : f(\mathbf{x}) \geq \theta \}$. This gives, 
\begin{flalign} \label{eq_th1_proof_a}
\mathbbm{P}_{\mathcal{D}_{\mathcal{T}|\mathcal{Y} = +1}} \{ \mathbf{x} \notin R_{f,1-\kappa}\} = \mathbbm{E}_{\mathcal{D}_{\mathcal{T}|\mathcal{Y} = +1}} [H(f(\mathbf{x}),1-\kappa)]
\end{flalign}
\noindent where, $H(x,\theta) = \left\{
\begin{array}{l l}
    0;\quad  \text{if} \; x \geq \theta  \\
    1;\quad  \text{else}  \\
\end{array}\right.$. For the rest of the proof we drop the subscripts as it is clear from context. To bound the R.H.S of \eqref{eq_th1_proof_a} we follow a similar approach of bounding a dominating function see Theorem 4.17 in \cite{shawe2004kernel}. Here we define,
\begin{flalign}
&A(x) = \left\{
\begin{array}{l l}
    0;\quad  \text{if} \; x > 1  \\
    \frac{1-x}{\kappa}; \quad  \text{if } \; 1-\kappa \leq x \leq 1  \\
    1 ; \quad \text{if } \; x < 1-\kappa \\
\end{array}\right.&& \nonumber
\end{flalign}
Note that, $A(x)$ is $\frac{1}{\kappa}-$ Lipchitz. Further, $H(f(\mathbf{x}), 1-\kappa) \leq A(f(\mathbf{x}))$. This gives,
$\mathbbm{E}[H(f(\mathbf{x}),1-\kappa)-1] \leq \mathbbm{E}[A(f(\mathbf{x}))-1] $.
Hence with probability $1-\eta, \forall f \in \mathcal{F}$ the following holds (see Theorem 4.9 in \cite{shawe2004kernel}); where $\mathbbm{\hat{E}} = $ the empirical estimate for the expectation operator.

\begin{flalign}
& \mathbbm{E}[H(f(\mathbf{x}),1-\kappa)-1]  \leq \mathbbm{\hat{E}}[A(f(\mathbf{x}))-1] && \nonumber \\
& \quad \quad \quad \quad \quad \quad \quad \quad \quad \quad + \mathcal{\hat{R}}_n((A-1) \circ \mathcal{F}) + 3\sqrt{\frac{ln \frac{2}{\eta}}{2n}}  \nonumber &&
\end{flalign}
From \text{Th. 4.15} \cite{shawe2004kernel}
\begin{flalign}
&\Rightarrow  \mathbbm{E}[H(f(\mathbf{x}),1-\kappa)]  \leq \quad  \frac{\sum\limits_{i=1}^n \xi_i}{\kappa n}  &&\nonumber \\
& \quad \quad \quad \quad \quad \quad  \quad \quad \quad  \quad \quad \frac{2}{\kappa}\mathcal{\hat{R}}_n(\mathcal{F}) + 3\sqrt{\frac{ln \frac{2}{\eta}}{2n}}   \quad  && \nonumber
\end{flalign}

where, $\xi_i = [1-f(\mathbf{x}_i)]_{+}$. Using \eqref{eq_th1_proof_a}, we get the final form of Theorem \eqref{th_oneclass}. \qed

\subsection{Proof of Theorem \ref{th_rademacher}} \label{sec:proof_th_rademacher}
\noindent \underline{Part (a):} It is clear that $\mathcal{F}_{\text{univ}} \subseteq \mathcal{F}_{\text{ind}}$. This ensures $\mathcal{\hat{R}}_n(\mathcal{F}_{\text{univ}}) \leq \mathcal{\hat{R}}_n(\mathcal{F}_{\text{ind}})$ (following Theorem 4.15 (i) in \cite{shawe2004kernel}. \qed \\

\noindent \underline{Part (b) - (i):} This follows from standard analysis (see Theorem 4.12 \cite{shawe2004kernel} or Lemma 22 in \cite{bartlett2002rademacher}). 
\begin{flalign}
&\hat{\mathcal{R}}_n(\mathcal{F}_{\text{ind}}) = \mathbbm{E}_{\sigma}[\underset{f \in \mathcal{F}}{\text{sup}} |\frac{2}{n} \sum_{i=1}^n \sigma_i f(\mathbf{x}_i)| \Big| (\mathbf{x}_i)_{i=1}^n]  && \nonumber  \\
&\quad \quad= \mathbbm{E}_{\sigma} [\underset{||\mathbf{w}||^2 \leq \Lambda^2}{\text{sup}} |\frac{2}{n} \sum_{i=1}^n \sigma_i f(\mathbf{x}_i)| \Big| (\mathbf{x}_i)_{i=1}^n]    && \nonumber \\
&\quad \quad = \mathbbm{E}_{\sigma} \big[\underset{||\mathbf{w}||^2 \leq \Lambda^2}{\text{sup}} |\frac{2}{n} \mathbf{w}^{\top} \big(\sum_{i=1}^n \sigma_i \phi(\mathbf{x}_i) \big)| \Big| (\mathbf{x}_i)_{i=1}^n \big]    && \nonumber \\
&\quad \quad \leq \mathbbm{E}_{\sigma} \big[\underset{||\mathbf{w}||^2 \leq \Lambda^2}{\text{sup}} \frac{2 ||\mathbf{w}||}{n}  || \big(\sum_{i=1}^n \sigma_i \phi(\mathbf{x}_i) \big) ||  \Big| (\mathbf{x}_i)_{i=1}^n \big]
&& \nonumber \\
&\quad \quad \leq \frac{2\Lambda}{n}\mathbbm{E}_{\sigma} \big[ || \sum_{i=1}^n \sigma_i \phi(\mathbf{x}_i)  ||  \Big| (\mathbf{x}_i)_{i=1}^n \big]    && \nonumber \\
&\quad \quad \leq \frac{2\Lambda}{n}\mathbbm{E}_{\sigma} \big[ || \sum_{i=1}^n \sigma_i \phi(\mathbf{x}_i)  ||^2  \Big| (\mathbf{x}_i)_{i=1}^n \big]^{\frac{1}{2}}  \quad \text{\small{(Jensen's inequality)}}  && \nonumber \\
&\quad \quad \leq \frac{2\Lambda}{n}\mathbbm{E}_{\sigma} \Big[  \sum_{i,j=1}^n \sigma_i \sigma_j \phi(\mathbf{x}_i) \phi(\mathbf{x}_j)   \big| (\mathbf{x}_i)_{i=1}^n \Big]^{\frac{1}{2}}    && \nonumber \\
&\quad \quad = \frac{2\Lambda}{n} \Big[\sum_{i=1}^n ||\phi(\mathbf{x}_i)||^2  \Big]^{\frac{1}{2}}    && \nonumber
\end{flalign} \qed

\noindent \underline{\text{Part (b) - (ii)}}: Define 
\begin{flalign} \label{eq_rad_fusvm}
&\mathcal{W}_{\text{univ}} = \{\mathbf{w} \big| ||\mathbf{w}||^2 \leq \Lambda^2 ; |\mathbf{w}^{\top} \phi(\mathbf{x}^*) - 1| \leq \Delta ; \forall \mathbf{x}^* \in \mathcal{X}_{U}^* \} && \nonumber \\
&\subseteq \{\mathbf{w} \big| \; ||\mathbf{w}||^2 \leq \Lambda^2 ; |\mathbf{w}^{\top} \mathbf{u}_j - 1| \leq \Delta ; && \nonumber \\
& \quad \quad \quad \quad  \quad \quad  \forall \mathbf{u}_j = \phi(\mathbf{x}_j^*) \; ; \; \forall \; \mathbf{x}_j^* \in \mathcal{X}_{U}^*; \; j = 1 \ldots m \} && 
\end{flalign}
\noindent $\because$ the constraint on  all $\mathbf{x}^* \in \mathcal{X}_{U}^* \Rightarrow$ constraint on $m-$ samples. Now, let's analyze the constraint $|\mathbf{w}^{\top}\mathbf{u}_j - 1| \leq \Delta$. This implies,  $\mathbf{w}^{\top}\mathbf{u}_j - 1 \leq \Delta;  1- \mathbf{w}^{\top}\mathbf{u}_j \leq \Delta$ (simultaneously). However, only one of the constraint is active. Hence, we re-write the constraint as, $\forall j \; ; \; \begin{bmatrix}
           \mathbf{w}^{\top}\mathbf{u}_j \\
   		   \mathbf{w}^{\top}(-\mathbf{u}_j)  
\end{bmatrix} \leq \begin{bmatrix}
           \Delta + 1 \\
   		   \Delta - 1  
\end{bmatrix}$. \\ 
\noindent Next define a mapping where we concatenate the reflected space. i.e. $\psi:\phi(\mathbf{x}^*) \rightarrow \begin{bmatrix} \phi(\mathbf{x}^*)^\top \\ -\phi(\mathbf{x}^*)^\top \end{bmatrix}$ and rewrite $V = \psi\Big( [\phi(\mathbf{x}_j^*)]_{j=1}^m \Big) = \begin{bmatrix}
           \phi(\mathbf{x}_1^*)^\top \\
           \phi(\mathbf{x}_2^*)^\top \\
           \vdots \\
           \phi(\mathbf{x}_m^*)^\top \\
   		   -\phi(\mathbf{x}_1^*)^\top \\
           -\phi(\mathbf{x}_2^*)^\top \\
           \vdots \\
           -\phi(\mathbf{x}_m^*)^\top \\ 
\end{bmatrix}$. This can be compactly re-written as, $V = \begin{bmatrix}
           1\\
           -1
\end{bmatrix} \otimes \begin{bmatrix}
           (\mathbf{u}_1)^T\\
   		   \vdots \\
   		   (\mathbf{u}_{m})^T
\end{bmatrix}$. This results to the overall  constraint in \eqref{eq_rad_fusvm} to be, 
\begin{flalign}
\mathcal{W}_{\text{univ}} \subseteq \{\mathbf{w} \big| ||\mathbf{w}||^2 \leq \Lambda^2\;;\; [V\mathbf{w}]_j \leq \epsilon_j; j = 1 \ldots 2m \} \nonumber
\end{flalign}
\noindent where, $\epsilon = \begin{bmatrix} \begin{rcases}
\; \quad \Delta+1 \;\\
\quad \quad \vdots \\
\; \quad \Delta+1 \;  
\end{rcases} \rotatebox[origin=c]{90}{m-\text{times} }
\\ 
\begin{rcases}
\; \quad \Delta-1 \; \\
\quad \quad \vdots \\
\; \quad \Delta-1 \; 
\end{rcases} \rotatebox[origin=c]{90} {m-\text{times} } \\ \end{bmatrix}$. \\ In essence for each constraint in \eqref{eq_rad_fusvm} we create $2 \times$ the constraints for both the original and reflected space to take care of the absolute value. \\
\noindent Now, 
\begin{flalign} \label{eq_rad_fusvm_2}
&\mathcal{W}_{\text{univ}} \subseteq \{ \mathbf{w} \big| ||\mathbf{w}||^2 \leq \Lambda^2 ; [V\mathbf{w}]_j \leq \epsilon_j \; \forall j = 1 \ldots 2m \}&& \nonumber \\
& \quad \subseteq \{ \mathbf{w} \big| ||\mathbf{w}||^2 \leq \Lambda^2 ; (\mathbf{w}^{\top}V^{\top}V\mathbf{w}) \leq 2m[\Delta^2 + 1] \}&& 
\end{flalign}
\noindent The last line follows as the element-wise constraint is relaxed by $||\cdot||_2^2$ constraint.

\noindent Next, from \eqref{eq_rad_fusvm_2} and assuming a fixed mapping $\phi(\cdot)$, for the given training data $Z = \begin{bmatrix} (\mathbf{z}_1)^\top \\ \vdots \\ (\mathbf{z}_n)^\top \\ \end{bmatrix} =  \begin{bmatrix}\phi(\mathbf{x}_1)^\top \\ \vdots \\ \phi(\mathbf{x}_n)^\top \\ \end{bmatrix}$ we have,
\begin{flalign}
&\mathcal{\hat{R}}(\mathcal{F}_{\text{univ}}) \nonumber &&\\
&\; \overset{(a)}{=} \frac{2}{n} \mathbbm{E}_{\sigma} \big[\underset{\mathbf{w} \in \mathcal{W}_{\text{univ}}}{\text{sup}} \boldsymbol \sigma^{\top} (Z\mathbf{w}) \big] && \nonumber \\
& \; \leq \frac{2}{n} \mathbbm{E}_{\sigma} \big[\underset{ \scriptsize{\begin{array}{c}
    ||\mathbf{w}||^2 \leq \Lambda^2 \\  (\mathbf{w}^{\top}V^{\top}V\mathbf{w}) \leq 2m[\Delta^2 + 1]
 \end{array}}} {\text{sup}} \boldsymbol \sigma^{\top} (Z\mathbf{w}) \;\big] \quad \text{(from \eqref{eq_rad_fusvm_2})} && \nonumber
 \end{flalign}
 Hence $\forall \gamma \geq 0$ and $\Gamma = \Lambda^2+ 2\gamma m(\Delta^2 + 1)$ we have,
 \begin{flalign}
 & \; \overset{(b)}{\leq} \frac{2}{n} \mathbbm{E}_{\sigma} \big[\underset{ 
    ||\mathbf{w}||^2 + \gamma (\mathbf{w}^{\top}V^{\top}V\mathbf{w}) \leq  \Gamma
 } {\text{sup}} \boldsymbol \sigma^{\top} (Z\mathbf{w}) \;\big]  && \nonumber \\
& \; = \frac{2}{n} \mathbbm{E}_{\sigma} \big[\underset{ 
    \mathbf{w}^{\top}\big(\frac{I+\gamma V^{\top}V}{\Gamma}\big)\mathbf{w} \leq  1
 } {\text{sup}} \boldsymbol \sigma^{\top} (Z\mathbf{w}) \;\big]  && \label{eq_st_pnt} \\
& \; \overset{(c)}{=} \frac{2}{n} \mathbbm{E}_{\sigma} \Big[ || \big( \frac{I+\gamma V^{\top} V}{\Gamma} \big)^{-\frac{1}{2}}Z^{\top}\boldsymbol\sigma || \Big] && \nonumber \\
& \; \leq \frac{2}{n} \Big[ \mathbbm{E}_{\sigma} \big[ ||\big( \frac{I+\gamma V^{\top} V}{\Gamma} \big)^{-\frac{1}{2}}Z^{\top}\boldsymbol\sigma|| \big]^2 \Big]^{\frac{1}{2}} \; \text{(Jensen's inequality)} && \nonumber  \end{flalign}
\begin{flalign}
& \; \overset{(d)}{=} \frac{2}{n} \Big[ tr \big[ Z \big( \frac{I+\gamma V^{\top} V}{\Gamma}\big)^{-1} Z^{\top} \big] \Big]^{\frac{1}{2}} \; (tr:= \text{Trace})&& \nonumber \\
& \; = \frac{2}{n} \big[\Lambda^2+2\gamma m(\Delta^2+1)\big]^{\frac{1}{2}} \Big[ tr \big[ Z \big(I+\gamma V^{\top} V \big)^{-1} Z^{\top} \big] \Big]^{\frac{1}{2}} \; && \nonumber \\
& \overset{(e)}{=} \big(\frac{2}{n}\big) (\Lambda^2 + 2\gamma m(\Delta^2+1))^{\frac{1}{2}}\Big[tr(Z^{\top}Z)  &&\nonumber\\
&\quad \quad \quad \quad \quad \quad  - \gamma tr[ZV^{\top}(I+\gamma VV^{\top})^{-1}VZ^{\top}]\Big]^{\frac{1}{2}} \quad (\forall \gamma \geq 0) &&\nonumber \\
& =\frac{2\Lambda}{n} \sqrt{||\sum\limits_{i=1}^n \phi(\mathbf{x}_i)||^2}\Big[ \big( 1+\frac{2\gamma m(\Delta^2+1)}{\Lambda^2}\big) && \nonumber \\
& \quad \quad \quad \quad  \quad \quad \quad \quad \big[1- \frac{\gamma tr[(I+\gamma VV^{\top})^{-1}(VZ^{\top}ZV^{\top})]}{tr(Z^{\top}Z)}\big]\Big]^{\frac{1}{2}} &&\nonumber\\
&\Rightarrow \mathcal{\hat{R}}(\mathcal{F}_{\text{univ}})  \overset{(f)}{\leq}\frac{2\Lambda}{n} \sqrt{||\sum\limits_{i=1}^n \phi(\mathbf{x}_i)||^2}\Big[ \big( 1+\frac{2\gamma m(\Delta^2+1)}{\Lambda^2}\big) && \nonumber \\
& \quad \quad \quad \quad  \quad \quad \quad \quad \quad \quad \quad \quad \big[1- \frac{\gamma tr(VZ^{\top}ZV^{\top})}{tr(I+\gamma VV^{\top}) \; tr(Z^{\top}Z)}\big]\Big]^{\frac{1}{2}} &&\nonumber
\end{flalign}
\noindent The (in)-equalities follow,
\begin{enumerate}[label=(\alph*)]
    \item from symmetry $\mathbf{w} \in \mathcal{W}_{USVM} \Rightarrow -\mathbf{w} \in \mathcal{W}_{USVM}$. Hence we drop the absolute term from definition. Also for simplicity we drop the conditional term. This is clear from context.
    \item since the conditions  $\begin{array}{c}
    ||\mathbf{w}||^2 \leq \Lambda^2 \\  (\mathbf{w}^{\top}V^{\top}V\mathbf{w}) \leq 2m[\Delta^2 + 1]
 \end{array} \\ \Rightarrow ||\mathbf{w}||^2 + \gamma (\mathbf{w}^{\top}V^{\top}V\mathbf{w}) \leq  \Gamma \quad \forall \gamma \geq 0$.
    \item stationary point of the constraint. A similar approach was previously used in \cite{rosenberg2007rademacher}.
    \item since (Rademacher variables) are drawn uniformly over $\sigma \sim \{-1, +1\}$; we cancel the cross-terms $\sigma_i\sigma_j$ under expectation $\mathbbm{E}_{\sigma}$.
    \item using Sherman-Morrison-Woodbury formula.
    \item from the matrix inequality II in \cite{patel1979trace}.
\end{enumerate} \qed

\section{Reproducibility} 
\subsection{Network Architectures} \label{app_net_arch}
\subsubsection{LeNet Architecture for CIFAR-10 experiments} \label{app_lenet_cifar}

\begin{figure}[h] 
\centering
\includegraphics[width=0.45\textwidth]{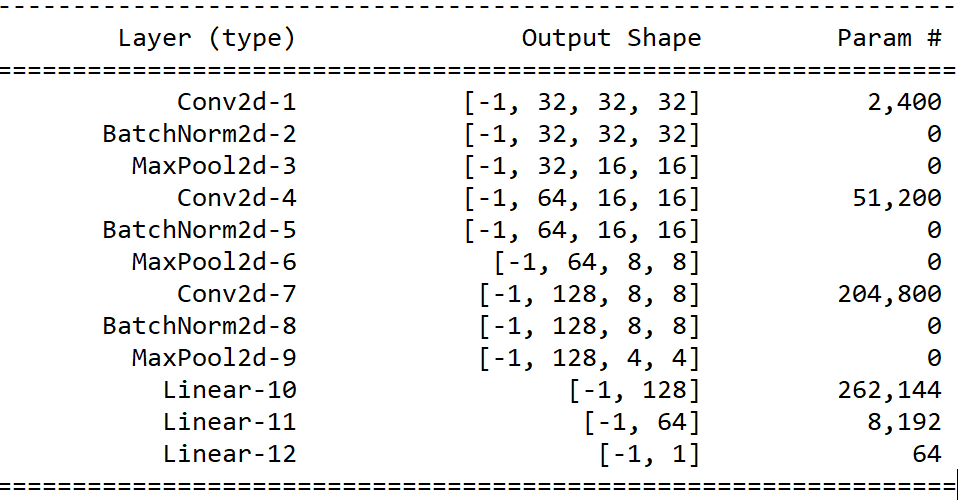}
\vskip -0.08in
\caption{Network used for CIFAR10 results in Table \ref{table:results1}}.
\label{fig_net_arc_cifar}
\vskip -0.1in
\end{figure}

For CIFAR-10 we use the same architecture (Fig. \ref{fig_net_arc_cifar}) as used in \cite{goyal2020drocc}.

\subsubsection{LeNet Architecture for MVTec experiments} \label{app_lenet_mvtec}
\begin{figure}[h] 
\centering
\includegraphics[height=0.30\textwidth]{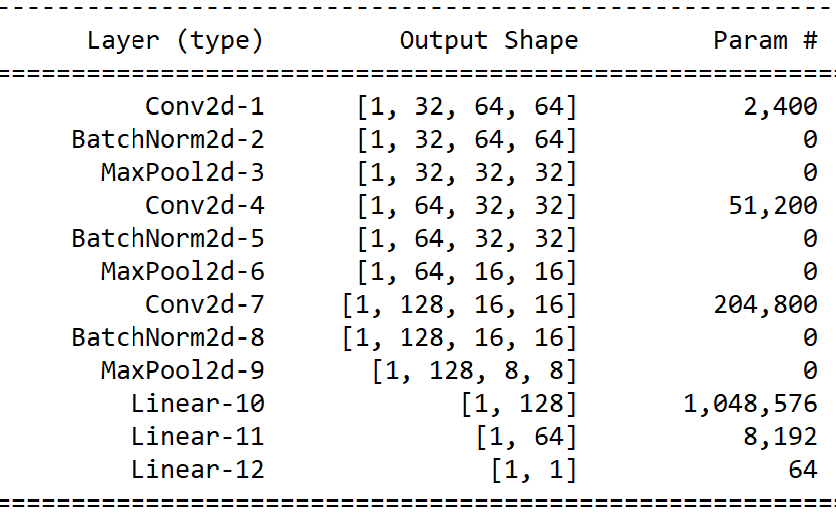}
\vskip -0.1in
\caption{Network used for MVTec-AD results in Table \ref{tab_MVtecAUC}}.
\label{fig_net_arc_mvtec}
\vskip -0.1in
\end{figure}

For MVTec-AD there have been a few recent works proposing advanced architectures to achieve state-of-the-art performance on this data \cite{carrara2020combining,huang2019attribute}. However, the main goal of our experiment is to illustrate the effectiveness of universum over inductive learning for one class problems. Hence, we stick to a simple LeNet architecture shown in Fig. \ref{fig_net_arc_mvtec}. 

Finally, for both the above architectures we use bias = False for convolution operations and set $\epsilon = 10^{-4}$, Affine = False for BatchNorm. Additionally, we use a leaky ReLU activation after every max-pool operation.

%\begin{table}[H]
%\caption{Network used for CIFAR10 results in Table \ref{table:results1}} \label{tab_net_arc_cifar}
%\vskip 0.15in
%\begin{center}
%\begin{small}
%\begin{sc}
%\begin{tabular}{rrr}
%\toprule
%Layer (type) & Output Shape & Param \# \\ \midrule
%Conv2d-1 & [-1, 32, 32, 32] & 2,400 \\
%BatchNorm2d-2 & [-1, 32, 32, 32] & 0 \\
%MaxPool2d-3 & [-1, 32, 16, 16] & 0 \\
%Conv2d-4 & [-1, 64, 16, 16] & 51,200 \\
%BatchNorm2d-5 & [-1, 64, 16, 16] & 0 \\
%MaxPool2d-6 & [-1, 64, 8, 8] & 0 \\
%Conv2d-7 & [-1, 128, 8, 8] & 204,800 \\
%BatchNorm2d-8 & [-1, 128, 8, 8] & 0 \\
%MaxPool2d-9 & [-1, 128, 4, 4] & 0 \\
%Linear-10 & [-1, 128] & 262,144 \\
%Linear-11 & [-1, 64] & 8,192 \\
%Linear-12 & [-1, 1] & 64 \\
%\bottomrule
%\end{tabular}
%\end{sc}
%\end{small}
%\end{center}
%\vskip -0.1in
%\end{table}

%\begin{table}[H]
%\caption{Network used for MVTEC-AD results in table \ref{tab_MVtecAUC}} \label{tab_net_arc_mvtec}
%\vskip 0.15in
%\begin{center}
%\begin{small}
%\begin{sc}
%\begin{tabular}{rrr}
%\toprule
%Layer (type) & Output Shape & Param \# \\ \midrule
%Conv2d-1 & [1, 32, 64, 64] & 2,400 \\
%BatchNorm2d-2 & [1, 32, 64, 64] & 0 \\
%MaxPool2d-3 & [1, 32, 32, 32] & 0 \\
%Conv2d-4 & [1, 64, 32, 32] & 51,200 \\
%BatchNorm2d-5 & [1, 64, 32, 32] & 0 \\
%MaxPool2d-6 & [1, 64, 16, 16] & 0 \\
%Conv2d-7 & [1, 128, 16, 16] & 204,800 \\
%BatchNorm2d-8 & [1, 128, 16, 16] & 0 \\
%MaxPool2d-9 & [1, 128, 8, 8] & 0 \\
%Linear-10 & [1, 128] & 1,048,576 \\
%Linear-11 & [1, 64] & 8,192 \\
%Linear-12 & [1, 1] & 64 \\
%\bottomrule
%\end{tabular}
%\end{sc}
%\end{small}
%\end{center}
%\vskip -0.1in
%\end{table}

\subsection{Model Parameters for Table \ref{table:results1}} \label{app_mod_sel_cifar}

\subsubsection{DOC and DOC$^3$ model parameters used in Table 1}
\begin{table}[h]
\caption{Optimal Model Parameters for CIFAR10 results for DOC.} \label{tab_cifarModParamDOC}
\vskip 0.15in
\begin{center}
\begin{small}
\begin{sc}
\begin{tabular}{c|cc}
\toprule
\specialcell{Class} & $\lambda$ & \specialcell{SGD\\(Learning Rate)} \\ \midrule
Airplane & 0.5 & 0.005 \\
Automobile & 1.0 & 0.001 \\
Bird & 0.5 & 0.005 \\
Cat & 1.0 & 0.001 \\
Deer & 1.0 & 0.001 \\
Dog & 0.5 & 0.001 \\
Frog & 0.5 & 0.001 \\
Horse & 1.0 & 0.001 \\
Ship & 0.5 & 0.001 \\
Truck & 0.001 & 0.001 \\
\bottomrule
\end{tabular}
\end{sc}
\end{small}
\end{center}
\vskip -0.1in
\end{table}

\begin{table}[h]
\caption{Optimal Model Parameters for CIFAR10 results for DOC$^3$.}\label{tab_cifarModParamDOC3}
\vskip 0.15in
\begin{center}
\begin{small}
\begin{sc}
\begin{tabular}{c|ccc}
\toprule
\specialcell{Class} & $\lambda$ & \specialcell{SGD\\(Learning Rate)}  & $ C_U/C$ \\ \midrule
Airplane & 0.1 & 0.005 & 0.5 \\
Automobile & 0.05 & 0.0005 & 0.1 \\
Bird & 0.05 & 0.005 & 0.5 \\
Cat & 0.1 & 0.005 & 1.0 \\
Deer & 0.1 & 0.001 & 1.0 \\
Dog & 0.05 & 0.001 & 1.0 \\
Frog & 0.1 & 0.005 & 0.5 \\
Horse & 0.05 & 0.0005 & 1.0 \\
Ship & 0.05 & 0.005 & 0.1 \\
Truck & 0.05 & 0.0001 & 0.05 \\
\bottomrule
\end{tabular}
\end{sc}
\end{small}
\end{center}
\vskip -0.1in
\end{table}

There are several hyper-parameters to be tuned for DOC and DOC$^3$. To simplify our analysis we fix a few of these parameters following prior research. 
\begin{itemize}
    \item Unlike previous works like \cite{ruff2018deep,goyal2020drocc}, we uniformly use an SGD optimizer with batch\_size = 256. Although, training for each class represent a completely different problem, we adopt this to maintain consistency and isolate out the effect of optimizers for DOC vs. DOC$^3$ performances.
    \item For DOC we fix the total number of iterations for gradient updates to 300. Except for class `DOG' and `Truck' we use 400 and 50 respectively. For DOC$^3$ we fix it to 350. This is in the same range as \cite{ruff2018deep}, and  hence incurs similar computation complexity as the baseline DOCC and DROCC algorithms.
    \item Finally for DOC$^3$ we fix $\Delta = 0$. 
\end{itemize}
With the above hyper parameters fixed our best selected remaining hyper parameters for DOC and DOC$^3$ are provided in Table \ref{tab_cifarModParamDOC} and \ref{tab_cifarModParamDOC3} respectively. 

\subsubsection{DOC (DA/OE) model parameters in Table 1}
Next, we provide the optimal model parameters for the DOC (DA/OE) setting in Table \ref{tab_Binary_DEEPSVM}. For the DOC (DA/OE) following \cite{goyal2020drocc} we introduce the universum samples as negative class in a standard binary hinge loss. The explicit form of this loss is also discussed in Appendix \ref{sec_synth} in eq. \eqref{eq_binSVM}. Here we set $C^{+} = C^{-} = 1$. 

\begin{table}[h]
\caption{DOC (DA/OE) parameters CIFAR-10.}  \label{tab_Binary_DEEPSVM}
\begin{center}
\begin{small}
\begin{sc}
\begin{tabular}{c|cc}
\toprule
\specialcell{CIFAR-10 \\ (Class)} & $\lambda = \frac{1}{2C} $ & \specialcell{SGD\\Learning Rate} \\ \midrule
Airplane & 1.0 &  $5 \times 10^{-3}$ \\
Automobile & 0.01& $5 \times 10^{-4}$ \\
Bird &0.5 & $5 \times 10^{-4}$\\
Cat & 1.0 & $10^{-4}$  \\
Deer & 1.0 & $10^{-4}$  \\
Dog & 0.5 & $10^{-3}$ \\
Frog & 0.5& $5\times 10^{-3}$  \\
Horse & 0.01 & $10^{-3}$ \\
Ship & 1.0 & $10^{-3}$ \\
Truck & 0.01 & $10^{-4}$ \\
\bottomrule
\end{tabular}
\end{sc}
\end{small}
\end{center}
\vskip -0.1in
\end{table}

\subsubsection{Model parameters and experiment set up for DROCC-LF under OE vs. Universum Setting} \label{app_drocc_lf}

For the DROCC-LF (OE) we use the same implementation as in \cite{goyal2020drocc}. For the DROCC-LF under universum setting we replace the binary cross entropy loss used in \cite{goyal2020drocc} with the universum loss in \eqref{eq_USVM} (see Algo \ref{alg:drocclf-nn}). Here we use the same notations as also used in \cite{goyal2020drocc}. Further, as in Section \ref{sec_mvtec_exp} we replace the relu operator $[x]_{+}$ with the softplus operator for the loss functions.   

We adopt the same LeNet architecture used in  \cite{ruff2018deep,goyal2020drocc} (see Fig \ref{fig_net_arc_cifar}). Finally, we run the experiments over 10 runs and report the best AUC over the range of parameters recommended in \cite{goyal2020drocc} (Section 5). That is learning rate = $10^{-4}$, radius ($r$) in range of $\sqrt{d}$ = $\{ 8.0, 16.0, 32.0\}$. Here, for both the methods we use Adam and fix the number of ascent steps = 10 and batch size = 256 and total epochs =  350. The remaining parameters are set to default values. The final optimal parameters selected for the different classes is provided in Table \ref{tab_DROCCLF_params}.

\begin{figure}[H]
    \begin{minipage}{.49\textwidth}
        \begin{algorithm}[H]
            \caption{DROCC-LF (under Universum setting)} \label{alg:drocclf-nn} \textbf{Input:} Training (normal) samples $\mathcal{T}=(\mathbf{x}_i,y_i = +1)_{i=1}^n$ and Universum samples $\mathcal{U} = (\mathbf{x}_{i^\prime}^{*})_{i^\prime=1}^m$.  
            
            \textbf{Parameters:} Radius $r$, $\lambda\geq 0$, $\mu\geq 0$, step-size $\eta$, number of gradient steps $m_g$, number of initial training steps $n_0$.
            
            \textbf{Initial steps:} For $B = 1, \hdots n_0$
            
            \quad Batch  of training ($X_T$) and  universum ($X_U$) samples
            
            \quad$\theta = \theta - \nabla\Big(\sum \limits_{\substack{\mathbf{x}_i \in X_B}} L_T(f(\mathbf{x}_i)) + \sum \limits_{\substack{\mathbf{x}_{i^\prime}^{*} \in X_U}} L_U(f(\mathbf{x}_{i^\prime}^{*}))  \Big)  $
            
            \textbf{DROCC steps:} For $B = n_0, \hdots n_0 + N$ 
            
             \quad $X_T$: Batch of normal training inputs ($y=+1$)
            
            %% \hspace*{\algorithmicindent} \hspace*{\algorithmicindent} \textbf{Adversarial search:}
            
           \quad $\forall x \in X_T: h \sim \mathcal{N}(0, I_{d})$
            %% ~~(Steps below in parallel across batch)

            \textbf{Adversarial search:} For $i = 1, \hdots m_g$
            
            \quad 1. $L_T(h) = L_T(f(x + h), -1)$
            %% Loss is high if $x+h$ is classified as positive
            
           \quad 2. $h = h + \eta \frac{\nabla_h L_T(h)}{\| \nabla_h L_T(h) \|}$
            %% Find point that has highest score of being positive
            
            \quad 3. $h = \text{Projection given by Prop.1 in \cite{goyal2020drocc}}$
            %% ~~Normalize gradient
            
           $\ell^{itr} =  \lambda\| \mathbf{w} \|^2 + \sum \limits_{\substack{\mathbf{x}_i \in X_B}} L_T(f(\mathbf{x}_i)) + \sum \limits_{\substack{\mathbf{x}_{i^\prime}^{*} \in X_U}} L_U(f(\mathbf{x}_{i^\prime}^{*})) +\mu  L_T(f(x + h), -1) $

            $\theta = \theta - \nabla \ell^{itr}$  
        \end{algorithm}
    \end{minipage}\hspace*{5pt}
    \begin{minipage}{.49\textwidth}

    \end{minipage}
\end{figure} 

\begin{table} 
\caption{DROCC (OE) and DROCC (Univ) optimal (radius) parameters used in Table \ref{table:results1} } \label{tab_DROCCLF_params}
%\vskip 0.15in
\begin{center}
\begin{small}
\begin{sc}
\begin{tabular}{c|cc}
\toprule
\specialcell{CIFAR-10 \\ (Class)} & \specialcell{DROCC-LF\\(OE)} & \specialcell{DROCC-LF\\(Univ)} \\ \midrule
Airplane & 8 &  8 \\
Automobile & 32& 8\\
Bird &16 & 8 \\
Cat &16 & 16  \\
Deer & 16 & 16  \\
Dog &16 & 8 \\
Frog & 32 &32  \\
Horse & 16 & 32 \\
Ship & 8 & 8 \\
Truck & 8 & 8 \\
\bottomrule
\end{tabular}
\end{sc}
\end{small}
\end{center}
\end{table}

\noindent \textbf{Caveat(s)}: We found a few caveats while running the DROCC-LF experiments. One major caveat is that the gradient ascent steps are prone to instabilities. Note that the DROCC-LF algorithm (Algo 2 in  \cite{goyal2020drocc}) scales the perturbation direction (h) by the norm of the gradient vector. This results to severe gradient explosion. Appropriate measures to alleviate this issue has to be taken. Another major caveat is that the additional gradient ascent updates results to high computation complexity. For example, for the experiments presented in this paper a typical DROCC-LF run (350 epoch) takes  $\sim 10^4$ secs compared to $\sim 10^3$ sec without the adversarial updates. The system configuration used here is,
\begin{itemize}[nosep]
    \item[--] CPU = AMD Ryzen 9 5950X 16 Core.
    \item[--] RAM = 32 GB.
    \item[--] GPU = NVIDIA GeForce RTX 3080
    \item[--] CUDA = 11.4
\end{itemize}

\subsection{Model Parameters for Table \ref{tab_MVtecAUC}} \label{app_mod_sel_mvtec}

\begin{table}[h]
\caption{Optimal Model Parameters for MVTEC-AD results.} \label{tab_MVtecModParam}
%\vskip 0.15in
\begin{center}
\begin{small}
\begin{sc}
\resizebox{0.46\textwidth}{!}{
\begin{tabular}{c|c|ccc}
\toprule
\multicolumn{2}{c|}{\specialcell{Method}} & $\lambda$ & \specialcell{Adam\\(Learning Rate)}  & $C_U/C$ \\ \midrule
\multicolumn{5}{c}{Leather} \\
\midrule
\multicolumn{2}{c|}{DOC} & 0.01 & $10^{-5}$& - \\ \cline{1-2}
\multirow{3}{*}{\rotatebox{90}{DOC$^3$}} & Noise & 0.01 & $10^{-5}$ & 0.01 \\
& Obj. & 0.01 & $10^{-5}$ & 0.1 \\
& Text. & 0.01 & $10^{-6}$ & 0.01 \\
\midrule
\multicolumn{5}{c}{Wood} \\
\midrule
\multicolumn{2}{c|}{DOC} & 0.005 & $10^{-4}$& - \\ \cline{1-2}
\multirow{3}{*}{\rotatebox{90}{DOC$^3$}} & Noise & 0.005 & $5\times10^{-6}$ & 1.0 \\
& Obj. & 0.005 & $10^{-5}$ & 1.0 \\
& Text. & 0.05 & $10^{-5}$ & 0.1 \\
\midrule
\multicolumn{5}{c}{Tile} \\
\midrule
\multicolumn{2}{c|}{DOC} & 0.01 & $10^{-5}$& - \\ \cline{1-2}
\multirow{3}{*}{\rotatebox{90}{DOC$^3$}} & Noise & 0.1 & $5\times10^{-6}$ & 0.1 \\
& Obj. & 0.005 & $10^{-4}$ & 2.0 \\
& Text. & 0.1 & $5\times10^{-6}$ & 0.1 \\
\midrule
\multicolumn{5}{c}{Carpet} \\
\midrule
\multicolumn{2}{c|}{DOC} & 1.0 & $5\times10^{-4}$& - \\ \cline{1-2}
\multirow{3}{*}{\rotatebox{90}{DOC$^3$}} & Noise & 0.001 & $10^{-5}$ & 0.01 \\
& Obj. & 0.005 & $5 \times 10^{-5}$ & 2.0 \\
& Text. & 1.0 & $10^{-3}$ & $10^{-5}$ \\
\bottomrule
\end{tabular}
}
\end{sc}
\end{small}
\end{center}
\vskip -0.1in
\end{table}

Here we provide the optimal model parameters selected and used to reproduce the DOC and DOC$^3$ results in Table \ref{tab_MVtecAUC} and \ref{tab_MVtecCorr}. For this set of experiments we  use the Adam optimizer with batch\_size = 100. Further, to simplify model selection we fix the total number of iterations to 1000, and $\Delta = 0$. Finally we also provide the optimal hyperparameters for the DOC (DA/OE) algorithm in Table \ref{tab_Binary_DEEPSVM_mvtec}.

\begin{table}[h]
\caption{DOC (DA/OE) model parameters for MVTEC results.} \label{tab_Binary_DEEPSVM_mvtec}
%\vskip 0.5in
\begin{center}
\begin{small}
\begin{sc}
\resizebox{0.46\textwidth}{!}{
\begin{tabular}{c|c|cc}
\toprule
%\rotatebox{90}{\scriptsize{+ class}} & - class & $\lambda = \frac{1}{2C} $ & Learning Rate \\ \midrule
Normal & \specialcell{Univ.\\Anomaly} & $\lambda = \frac{1}{2C} $ & Learning Rate \\ \midrule
\midrule
\multirow{3}{*}{\rotatebox{90}{\scriptsize{Leather}}} & Noise & 0.001 &  $10^{-6}$ \\
& Obj. & 0.01 &  $10^{-5}$ \\
& Text. & 0.0001 &  $10^{-5}$ \\
\midrule
\multirow{3}{*}{\rotatebox{90}{Wood}} & Noise & 0.01 &  $10^{-5}$ \\
& Obj. & 0.01 &  $10^{-6}$ \\
& Text. & 0.01 &  $10^{-6}$ \\
\midrule
\multirow{3}{*}{\rotatebox{90}{Tile}} & Noise & 0.005 &  $10^{-6}$ \\
& Obj. & 0.001 &  $10^{-4}$ \\
& Text. & 0.1 &  $10^{-6}$ \\
\midrule
\multirow{3}{*}{\rotatebox{90}{\scriptsize{Carpet}}} & Noise & 0.001 &  $10^{-6}$ \\
& Obj. & 0.01 &  $10^{-5}$ \\
& Text. & 0.005 &  $10^{-5}$ \\
\bottomrule
\end{tabular}
}
\end{sc}
\end{small}
\end{center}
\end{table}

\subsection{Ablation Study Hyperparameters}
The DOC$^3$ algorithm mainly introduces two additional hyper-parameters $C_U$ and $\Delta$ compared to its inductive counterpart. The success of such an advanced technique depends on careful tuning of the hyperparameters. In this section we perform an ablation study of the $\frac{C_U}{C}$ and the $\Delta$ hyperparameters. To simplify we present the results for the CIFAR10 data. Analysis using MVTec-AD data provides similar conclusions. 

\begin{figure*}[!htb]%
\centering
\subfigure[Airplane]{\includegraphics[width=.2\linewidth]{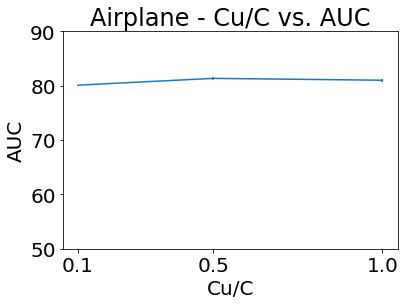}}
\subfigure[Automobile]{\includegraphics[width=.2\linewidth]{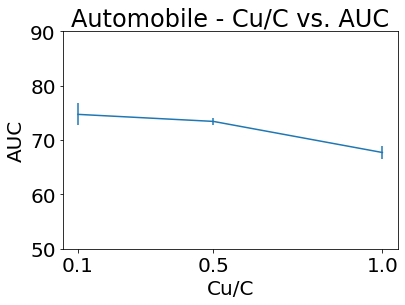}}
\subfigure[Bird]{\includegraphics[width=.2\linewidth]{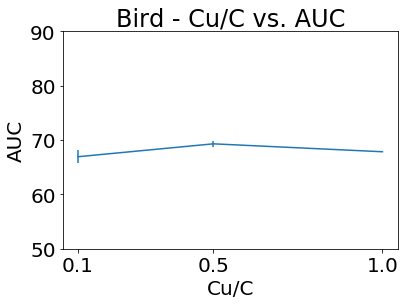}} 
\subfigure[Cat]{\includegraphics[width=.2\linewidth]{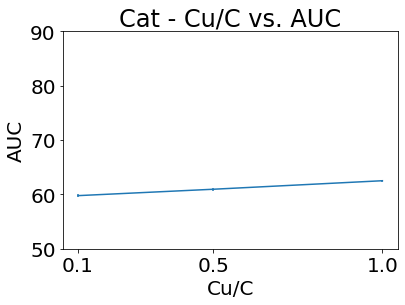}} 
\subfigure[Deer]{\includegraphics[width=.2\linewidth]{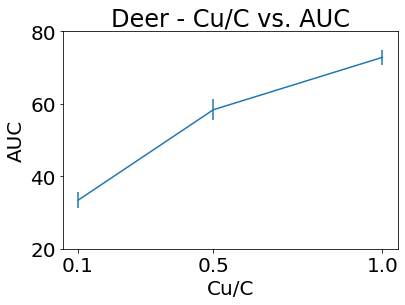}}
\subfigure[Dog]{\includegraphics[width=.2\linewidth]{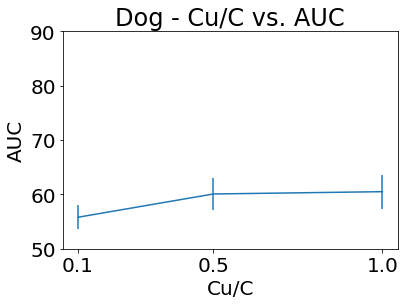}} 
\subfigure[Frog]{\includegraphics[width=.2\linewidth]{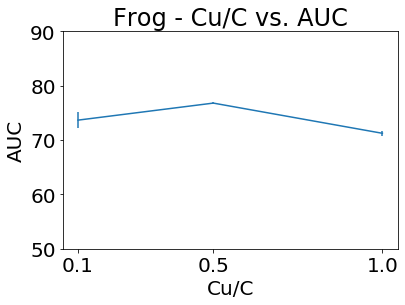}} 
\subfigure[Horse]{\includegraphics[width=.2\linewidth]{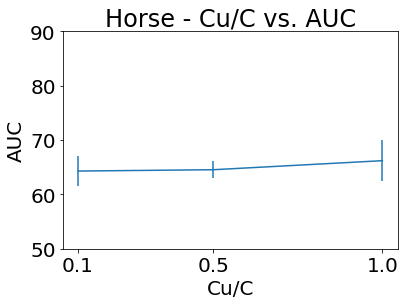}}
\subfigure[Ship]{\includegraphics[width=.2\linewidth]{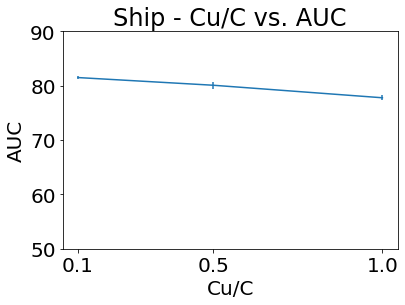}}
\subfigure[Truck]{\includegraphics[width=.2\linewidth]{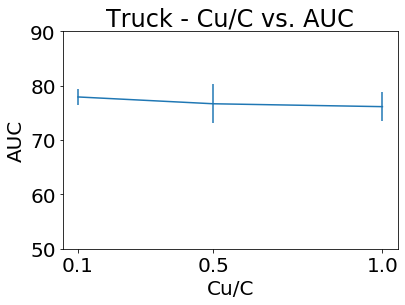}}
\caption{Ablation study - AUC values for varying $C_U/C$ values.}
\label{ablation_Cu}
\end{figure*}

\begin{figure*}[!htb]%
\centering
\subfigure[Airplane]{\includegraphics[width=.2\linewidth]{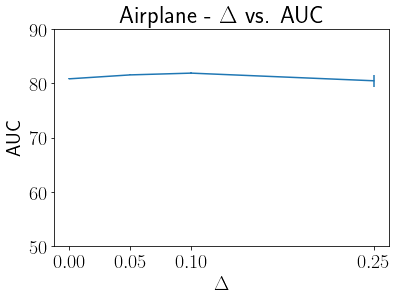}}
\subfigure[Automobile]{\includegraphics[width=.2\linewidth]{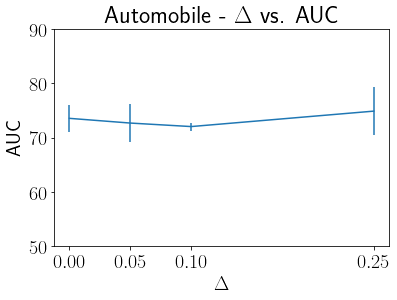}}
\subfigure[Bird]{\includegraphics[width=.2\linewidth]{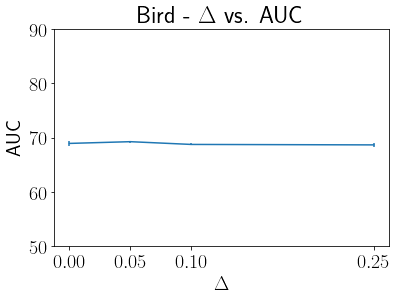}} 
\subfigure[Cat]{\includegraphics[width=.2\linewidth]{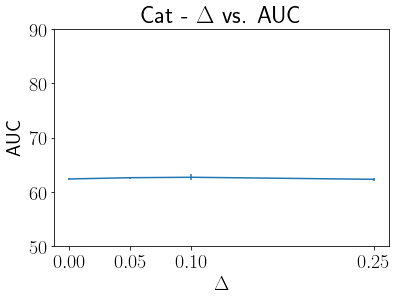}} 
\subfigure[Deer]{\includegraphics[width=.2\linewidth]{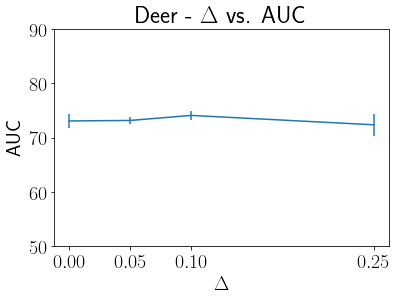}}
\subfigure[Dog]{\includegraphics[width=.2\linewidth]{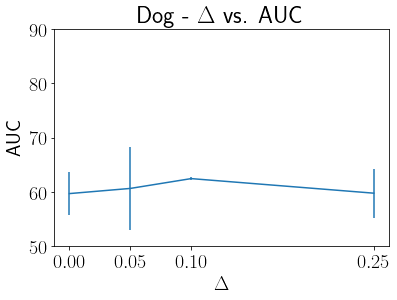}} 
\subfigure[Frog]{\includegraphics[width=.2\linewidth]{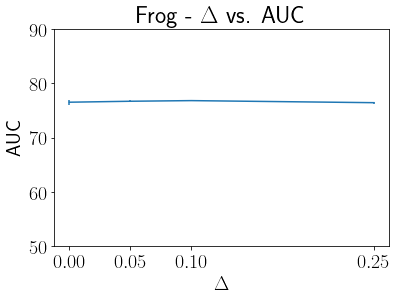}} 
\subfigure[Horse]{\includegraphics[width=.2\linewidth]{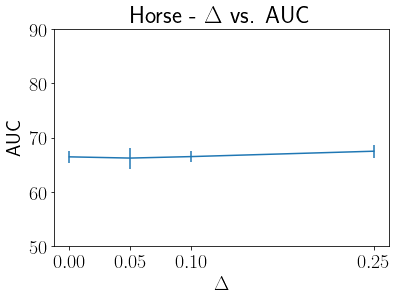}}
\subfigure[Ship]{\includegraphics[width=.2\linewidth]{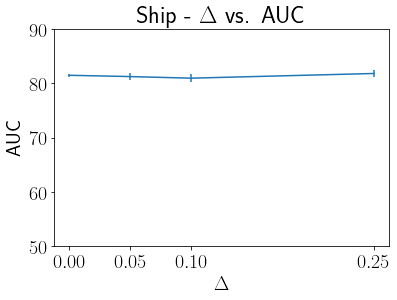}}
\subfigure[Truck]{\includegraphics[width=.2\linewidth]{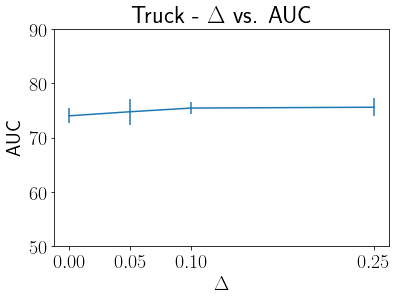}}
\caption{Ablation study - AUC values for varying $\Delta$ values.}
\label{ablation_delta}
\end{figure*}

Figs \ref{ablation_Cu} and \ref{ablation_delta} provides the average $\pm$ std. deviation of the AUC values over 10 experiment runs for varying $\frac{C_U}{C}$ - ratios and $\Delta$ values respectively. The experiment follows the same setting as in Section \ref{sec_mvtec_exp}. Further all the other model parameters are set to their optimal values reported in Table \ref{tab_MVtecModParam}. As seen from the figures, the model performance significantly varies for different $\frac{C_U}{C}$ -values (specifically for automobile, deer, dog, frog etc.). On the contrary, the DOC$^3$ model performance seems relatively stable for varying $\Delta$ values (see Fig \ref{ablation_delta}). Such behavior is also seen for the MVTec-AD dataset. In line with this analysis throughout the paper we fix $\Delta = 0$ and follow the current norm of reporting the best model’s results over a small subset of hyperparameters. But this is far from practical. This motivates advanced mechanisms for optimal selection of this hyper parameter, which is still an open research topic. From our prior experiments, we found $C_U/C$ in the range of $[0.01, 1.0]$ provides reasonable performance in practice.

\section{Additional Experiments and Results} \label{sec:tables}
\subsection{Comparisons of Disjoint Auxiliary (or Outlier Exposure) vs. Universum settings} \label{sec:OEcomparison}

In this section we highlight the differences between the universum vs. the `Disjoint Auxiliary data' setting used in \cite{dhar2014analysis} (see Section 4.3) and  \cite{hendrycks2018deep,goyal2020drocc} etc. As discussed in the Section \ref{sect:related_works} a major difference is the assumption  that the universum samples act as contradictions to the unseen anomalous class (see Definition \eqref{def_univsetting}). Methods using the `Disjoint Auxiliary' setting do not use this assumption and formulate a loss function which only contradicts the `normal' class. Such approaches have also been called `Supervised OE' in \cite{ruff2021unifying} or `Limited Negatives' in \cite{goyal2020drocc}. In this section we take a more pedagogical approach to highlight the differences between Universum vs. `Disjoint Auxiliary' setting. For simplicity we use a binary classifier as an exemplar of this `Disjoint Auxiliary' setting. That is, we build a binary classifier with `$+1$' (normal samples) and `$-1$' (contradiction a.k.a universum) samples. Note that, such an approach is philosophically inconsistent following Def. \ref{def_univsetting}; where the universum samples are assumed to not follow the same distribution as both the normal (`+1') and anomalous (`-1') class. Using the universum samples as (`-1') class violates the assumption that universum follows a different distribution than the anomalous class. To further confirm our theoretical analysis we provide a simple synthetic example in \ref{sec_synth}. Empirical comparisons using the CIFAR-10 and MV-Tec AD datasets are already provided in the main text in Tables \ref{table:results1} and \ref{tab_MVtecAUC} respectively. We further consolidate our claim by comparing the adversarial setting based DROCC-LF (extended under OE setting) \cite{goyal2020drocc}  vs. DROCC-LF (extended under universum setting) in Table \ref{table:results1} (also discussed in section \ref{app_drocc_lf})

\subsubsection{Synthetic Experiment} \label{sec_synth}
\begin{figure}[h]
\centering
\includegraphics[height=9.0cm, width=5.5cm]{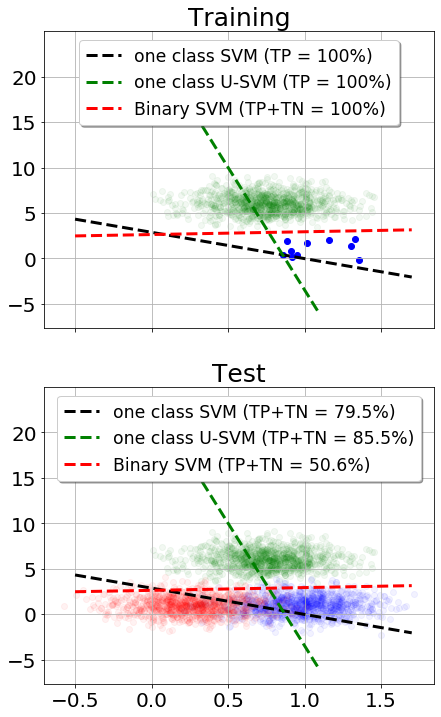}
\caption{Decision boundaries for \textbf{one class SVM} vs. \textcolor{darkgreen}{one class U-SVM} vs. \textcolor{red}{binary SVM}. Typical parameters, \textbf{one class SVM} ($C$ = 5), 
\textcolor{darkgreen}{one class U-SVM} ($C_U = 10^{-3}, \Delta = 0$), 
\textcolor{red}{binary SVM} ($C$ = 10). TP = True Positive, TN = True Negatives. } \label{fig_simpleExample} 
\end{figure}

For our synthetic example, we use synthetic data generated using normal distribution $\mathcal{N}(\mu,\sigma)$. For illustration we use,

\begin{itemize}[nosep]
    \item Normal Class (+1) : $\mu = (1.0, 1.0)$, $\sigma = (0.25,1.0)$.
    \item Anomaly Class (-1) : $\mu = (0.25, 1.0)$, $\sigma = (0.25,1.0)$.
    \item Contradictions : $\mu = (0.75, 6.0)$, $\sigma = (0.25,1.0)$.
\end{itemize}

Additionally we use,

\begin{itemize}[nosep]
    \item No. of Training samples (+1 class) = 10
    \item No. of Test samples (+1, -1) class = 1000 per class.
    \item No. of Universum samples = 1000.
\end{itemize}

Note, that in the above synthetic example the discriminative power is mostly contained in the 1$^{st}$ dimension. Having `good' universum samples can incorporate this additional information by contradicting the 2$^{nd}$ dimension while estimating the decision rule. This is also seen from Fig. \ref{fig_simpleExample}. Fig. \ref{fig_simpleExample} provides the decision boundaries obtained under inductive \eqref{eq_SVM} vs. universum settings \eqref{eq_USVM} using only linear parameterization. Under linear parameterization the formulations reduces to standard SVM formulations so we refer them as one class SVM and one class U-SVM respectively. Finally, we also provide the decision boundary using a binary SVM \cite{cherkassky2007learning}. 

\begin{flalign} \label{eq_binSVM}
& \underset{\mathbf{w}}{\text{min}}\quad \frac{1}{2}||\mathbf{w}||_2^2 + C \;  \sum_{i=1}^n \Big[ C^+[1-\mathbf{w}^\top\phi(\mathbf{x}_i)]_+ && \\
& \quad \quad \quad \quad  \quad \quad \quad \quad \quad \quad + \quad  C^-[1+\mathbf{w}^\top\phi(\mathbf{x}_i)]_+ \Big] && \nonumber
\end{flalign}

\noindent where, $[x]_+ = \text{max}(0,x) $. For the binary SVM we use the universum as (-1) class, and adopt cost-sensitive formulation with a cost ratio $\frac{C^+}{C^-} = \frac{\# univ}{\# train} = \frac{1000}{10}$, to handle the class imbalance.

As seen from Fig. \ref{fig_simpleExample}, using binary formulation in this universum setting does not correctly capture information available through the contradiction samples. That is, discriminating between normal and contradiction samples does not provide a good classifier for normal vs. anomaly classification. The one class SVM although correctly classifies the positive samples (TP = 100\%); does not perform good on future test samples. Using the universum samples, we can incorporate the additional information that the decision boundary should align along the vertical axis to have maximal contradiction (following Prop. \ref{prop_maxcontradiction}). And by doing so, it improves the test performance over the one class SVM solution.

\subsection{Reproducing DOCC \cite{ruff2018deep} and DROCC \cite{goyal2020drocc}  Results} \label{app_docc_drocc_reproduce}
\subsubsection{Deep One Class Classification (DOCC) Results}

%\vskip -0.3in
\begin{table}[h]
\caption{Reproducing DOCC results in \cite{ruff2018deep}} \label{tab_cifarDOCCRep_Ruff}
%\vskip 0.15in
\begin{center}
\resizebox{\columnwidth}{!}{
\begin{small}
\begin{sc}
\begin{tabular}{c|c|cc}
\toprule
%\specialcell{Class} & Paper & \multicolumn{2}{c}{\specialcell{DOCC\\$\nu=0.1$ & $\nu=0.1$}} & \multicolumn{2}{c}{\specialcell{DOCC (soft-boundary)\\$\nu=0.1$ $\nu=0.1$}} \\ \midrule
 & & \multicolumn{2}{c}{DOCC (our Run)} \\
\specialcell{Class} & \specialcell{Ruff et al.,\\ 2018 \cite{ruff2018deep}} & $\nu=0.1$  & $\nu=0.01$ \\ \midrule
%\specialcell{Class} & Paper & \specialcell{DOCC\\$\nu=0.1$}  & \specialcell{DOCC\\$\nu=0.01$} & \specialcell{DOCC (soft-boundary)\\$\nu=0.1$} & \specialcell{DOCC (soft-boundary)\\$\nu=0.01$} \\ \midrule
Airplane & $61.7\pm4.1$ & $61.0\pm1.6$ & $61.1\pm1.4$ \\
Automobile & $65.9\pm2.1$ & $60.4\pm1.7$ & $60.4\pm1.7$ \\
Bird & $50.8\pm0.8$ & $48.6\pm0.6$ & $48.6\pm0.6$ \\
Cat & $59.1\pm1.4$ & $57.7\pm0.9$ & $57.8\pm1.0$ \\
Deer & $60.9\pm1.1$ & $56.2\pm0.7$ & $56.3\pm0.8$ \\
Dog & $65.7\pm2.5$ & $63.4\pm1.1$ & $63.5\pm1.3$ \\
Frog & $67.7\pm2.6$ & $56.8\pm1.8$ & $56.9\pm2.0$ \\
Horse & $67.3\pm0.9$ & $59.9\pm1.9$ & $59.8\pm1.9$ \\
Ship & $75.9\pm1.2$ & $77.1\pm1.0$ & $76.9\pm1.0$ \\
Truck & $73.1\pm1.2$ & $66.9\pm0.6$ & $66.9\pm0.7$ \\
\bottomrule
\end{tabular}
\end{sc}
\end{small}}
\end{center}
%\vskip -0.3in
\end{table}

For the DOCC results we see very similar results for our run except the `Frog' and `Dog' classes; where the difference are not too significant. Hence, we report the results as presented in the paper.

\begin{table}[h]
\caption{Reproducing DOCC (soft-boundary) results in \cite{ruff2018deep}}\label{tab_cifarDOCCSoftRep}
\begin{center}
\resizebox{\columnwidth}{!}{
%\vskip 0.15in

\begin{small}
\begin{sc}
\begin{tabular}{c|c|cc}
\toprule
%\specialcell{Class} & Paper & \multicolumn{2}{c}{\specialcell{DOCC\\$\nu=0.1$ & $\nu=0.1$}} & \multicolumn{2}{c}{\specialcell{DOCC (soft-boundary)\\$\nu=0.1$ $\nu=0.1$}} \\ \midrule
 & & \multicolumn{2}{c}{\specialcell{DOCC soft-boundary\\(Our Run)}} \\
\specialcell{Class} & \specialcell{Ruff et al.,\\ 2018 \cite{ruff2018deep}} & $\nu=0.1$  & $\nu=0.01$ \\ \midrule
%\specialcell{Class} & Paper & \specialcell{DOCC\\$\nu=0.1$}  & \specialcell{DOCC\\$\nu=0.01$} & \specialcell{DOCC (soft-boundary)\\$\nu=0.1$} & \specialcell{DOCC (soft-boundary)\\$\nu=0.01$} \\ \midrule
Airplane & $61.7\pm4.1$ & $61.9\pm1.6$ & $62.5\pm1.9$ \\
Automobile & $65.9\pm2.1$ & $61.6\pm1.7$ & $62.6\pm2.0$ \\
Bird & $50.8\pm0.8$ & $48.0\pm0.9$ & $45.9\pm1.6$ \\
Cat & $59.1\pm1.4$ & $56.6\pm1.3$ & $56.0\pm1.8$ \\
Deer & $60.9\pm1.1$ & $56.2\pm0.8$ & $56.1\pm1.2$ \\
Dog & $65.7\pm2.5$ & $61.9\pm0.9$ & $60.7\pm1.7$ \\
Frog & $67.7\pm2.6$ & $59.8\pm1.8$ & $61.0\pm1.5$ \\
Horse & $67.3\pm0.9$ & $61.5\pm1.6$ & $61.3\pm1.5$ \\
Ship & $75.9\pm1.2$ & $77.7\pm0.9$ & $76.7\pm0.8$ \\
Truck & $73.1\pm1.2$ & $67.5\pm0.9$ & $68.7\pm1.4$ \\
\bottomrule
\end{tabular}
\end{sc}
\end{small}}
\end{center}
\vskip -0.1in
\end{table}

\subsubsection{Deep Robust One Class Classification (DROCC) Results}
% \subsection{Caveats with Deep Learning based One-Class Classification - the Maximum Data Piling Effect} \label{app_mdp}

% ORIGINAL TABLE

%\begin{table}[H]
%\caption{Deep Robust One Class Classification replication results} %\label{tab_cifarDOCCRep}
%\vskip 0.15in
%\begin{center}
%\begin{small}
%\begin{sc}
%% \tabcolsep = 0.1cm
%\begin{tabular}{c|ccc}
%\toprule
%\specialcell{Class} & Paper & \specialcell{DROCC\\(all-class scale)}  & %\specialcell{DROCC\\(no-prior scale)} \\ \midrule
%Airplane & $81.66\pm0.22$ & $79.99\pm1.65$ & $79.24\pm1.95$ \\
%Automobile & $76.73\pm0.99$ & $74.61\pm2.57$ & $74.92\pm2.66$ \\
%Bird & $66.66\pm0.96$ & $69.56\pm0.94$ & $68.29\pm1.53$ \\
%Cat & $67.13\pm1.51$ & $54.54\pm3.71$ & $62.25\pm2.67$ \\
%Deer & $73.62\pm2.00$ & $65.85\pm2.94$ & $70.34\pm2.68$ \\
%Dog & $74.43\pm1.95$ & $66.47\pm3.16$ & $66.18\pm2.09$ \\
%Frog & $74.42\pm0.92$ & $70.64\pm2.40$ & $68.16\pm2.12$ \\
%Horse & $71.39\pm0.22$ & $70.18\pm2.42$ & $71.33\pm4.57$ \\
%Ship & $80.01\pm1.69$ & $63.58\pm7.88$ & $62.39\pm10.33$ \\
%Truck & $76.21\pm0.67$ & $75.12\pm1.92$ & $76.58\pm1.94$ \\
%\bottomrule
%\end{tabular}
%\end{sc}
%\end{small}
%\end{center}
%\vskip -0.1in
%\end{table}

Here, we report the results of our run with two different scaling. For the `all-class' scale we use the scale used in the original DROCC paper \cite{goyal2020drocc} i.e. $\mu = (0.4914,0.4822, 0.4465)$ and standard deviation, $\sigma= (0.247, 0.243, 0.261)$. Note that, this scale is computed using the pixel values for all the classes. This in general is not available during training a one class classifier. Alternatively, `no-prior' scale also reports the results using a scale using $ \mu = (0.5, 0.5, 0.5)$ and standard deviation, $\sigma=(0.5, 0.5, 0.5)$. This scale does not need additional information from the other class's pixel values. We do not see a significant difference using these different scales. Although our re-runs show a significant difference for the `ship' class between our results and the paper. We report the results of our re-run using the `no-prior' scale in Table \ref{table:results1}.

\begin{table}[h]
\caption{Reproducing DROCC results in \cite{goyal2020drocc}} \label{tab_cifarDOCCRep_Goyal}
%\vskip 0.15in
\begin{center}
\begin{small}
\begin{sc}
\resizebox{0.5\textwidth}{!}{
\begin{tabular}{c|ccc}
\toprule
\specialcell{Class} & \specialcell{Goyal et al.,\\ 2020 \cite{goyal2020drocc}} & \specialcell{DROCC\\(all-class\\scale)}  & \specialcell{DROCC\\(no-prior\\scale)} \\ \midrule
Airplane & $81.66\pm0.22$ & $79.99\pm1.65$ & $79.24\pm1.95$ \\
Automobile & $76.73\pm0.99$ & $74.61\pm2.57$ & $74.92\pm2.66$ \\
Bird & $66.66\pm0.96$ & $69.56\pm0.94$ & $68.29\pm1.53$ \\
Cat & $67.13\pm1.51$ & $54.54\pm3.71$ & $62.25\pm2.67$ \\
Deer & $73.62\pm2.00$ & $65.85\pm2.94$ & $70.34\pm2.68$ \\
Dog & $74.43\pm1.95$ & $66.47\pm3.16$ & $66.18\pm2.09$ \\
Frog & $74.42\pm0.92$ & $70.64\pm2.40$ & $68.16\pm2.12$ \\
Horse & $71.39\pm0.22$ & $70.18\pm2.42$ & $71.33\pm4.57$ \\
Ship & $80.01\pm1.69$ & $63.58\pm7.88$ & $62.39\pm10.33$ \\
Truck & $76.21\pm0.67$ & $75.12\pm1.92$ & $76.58\pm1.94$ \\
\bottomrule
\end{tabular}
}
\end{sc}
\end{small}
\end{center}
\vskip -0.1in
\end{table}

\subsection{Additional Results on Tabular data from \cite{goyal2020drocc}} \label{app_tab_results}

In this section we provide additional results on the tabular data used in \cite{goyal2020drocc}. The data used involves standard anomaly detection problem described next,
\begin{itemize}[nosep]
    \item \textbf{Abalone} as used in \cite{das2018active} : Here the task is to predict the age of abalone using several physical measurements like, rings, sex, length, diameter, height, weight, etc. For this problem class 3 and 21 are anomalies and class 8, 9 and 10 serve as normal samples.
    \item \textbf{Arrhythmia} as used in \cite{zong2018deep}. Here the task is to identify the arrhythmic samples using the ECG features. We follow the same data set preparation as \cite{zong2018deep}. 
    \item \textbf{Thyroid} as used in \cite{zong2018deep}. The goal is to predict if a patient is hypothyroid based on his/her medical history.We follow the same data set preparation as \cite{zong2018deep}. 
\end{itemize}
\noindent For all the above data we use the data set preparation codes provided in \cite{goyal2020drocc}. This code provides the data preprocessing and partitioning scheme as used in the previous works. 

We follow the same experiment setup and network architecture  as in \cite{goyal2020drocc}. Table \ref{app_tab_tabular_small} provides the results of DOC$^3$ over 10 random partition of the data set. In each partition, we create training/test data as used in \cite{goyal2020drocc}. Note however, different from  \cite{goyal2020drocc}, we scale the data in the range of $[-1, +1]$ uniformly. In addition, here we generate uniform noise in range $[-1,+1]$ and use that as universum/contradiction samples. As seen from Table \ref{app_tab_tabular_small} the DOC$^3$ outperforms all existing approaches (except DROCC) for the Thyroid data; and significantly improves ($>$ 5 - 15 \%) upon the state-of-the-art results for the Arrhythmia and Abalone data. The optimal model parameters used for the results is also provided in Table \ref{tab_tabularParamDOC3} for reproducibility.

\begin{table}[H]
\centering
\caption{F1-Score $\pm$ standard deviation for one-vs-all anomaly detection on Thyroid, Arrhythmia, and Abalone datasets.} \label{app_tab_tabular_small}
\resizebox{\columnwidth}{!}{
\begin{tabular}{llll}
\hline
\multicolumn{1}{c}{\textbf{}} & \multicolumn{3}{c}{\textbf{F1-Score}}                                                   \\ \cline{2-4} 
\textbf{Method}               & \textbf{Thyroid}            & \textbf{Arrhythmia}         & \textbf{Abalone}            \\ \hline
OC-SVM    \cite{oneclasssvm}                    & 0.39 $\pm$ 0.01          & 0.46 $\pm$ 0.00         & 0.48 $\pm$ 0.00         \\
DCN\cite{caron2018deep}                           & 0.33 $\pm$ 0.03          & 0.38 $\pm$ 0.03          & 0.40 $\pm$ 0.01         \\
E2E-AE  \cite{zong2018deep}                      & 0.13 $\pm$ 0.04          & 0.45 $\pm$ 0.03          & 0.33 $\pm$ 0.03          \\
LOF     \cite{breunig2000lof}                      & 0.54 $\pm$ 0.01          & 0.51 $\pm$ 0.01          & 0.33 $\pm$ 0.01          \\
DAGMM \cite{zong2018deep}                        & 0.49 $\pm$ 0.04          & 0.49 $\pm$ 0.03          & 0.20 $\pm$ 0.03          \\
DeepSVDD \cite{ruff2018deep}           & 0.73 $\pm$ 0.00          & 0.54 $\pm$ 0.01          & 0.62 $\pm$ 0.01          \\
GOAD \cite{bergman2020classification}           & 0.72 $\pm$ 0.01          & 0.51 $\pm$ 0.02          & 0.61 $\pm$ 0.02          \\
DROCC \cite{goyal2020drocc}  & \textbf{0.78 $\pm$ 0.03} & 0.69 $\pm$ 0.02 & 0.68 $\pm$ 0.02 \\
DOC$^3$ (ours)  & 0.74 $\pm$ 0.01 & \textbf{0.73 $\pm$ 0.01} & \textbf{0.77 $\pm$ 0.01} \\
\hline
\end{tabular}
}
\label{tab:tabular}
\vspace{5pt}
\end{table}

\begin{table}[h]
\caption{Optimal Model Parameters for the Tabular data sets.}\label{tab_tabularParamDOC3}
\vskip 0.15in
\begin{center}
\resizebox{\columnwidth}{!}{
\begin{small}

\begin{tabular}{c|ccccc}
\toprule
\specialcell{Class} & $\lambda$ & \specialcell{Learning\\Rate}  & $ C_U/C$ & Epoch & Batch Size \\ \midrule
Thyroid & $10^{-6}$ & $10^{-3}$ & 5.0 & 500 & 100 \\
Arrhythmia & 0.01 & $10^{-3}$ & 0.001 & 300 & 100 \\
Abalone & 0.1 & $10^{-3}$ & 0.01& 300 & 100 \\
\bottomrule
\end{tabular}
\end{small}}
\end{center}
\vskip -0.1in
\end{table}

\section{Future Research} \label{sec_futureresearch}
Broadly there are two major future research directions,

\textbf{Model Selection} This is a generic issue for any (unsupervised) one class based anomaly detection formulation, and is further complicated by the non-convex loss landscape for deep learning problems. For DOC$^3$ we simplify model selection by fixing $\Delta = 0$, and optimally tuning $C_U$. However, the success of DOC$^3$ heavily depends on carefully tuning of its hyperparameters. In the absence of any validation set containing both `normal' and `anomalous' samples, we follow the current norm of reporting the best model's results over a small subset of hyperparameters. But this is far from practical. We believe, our Theorem \ref{th_oneclass} provides a good framework for bound based model selection. This in conjunction with Theorem \ref{th_rademacher} and the recent works on ERC for deep architectures \cite{neyshabur2015norm,sokolic2016lessons}, may provide better mechanisms for model selection and yield optimal models. 

\textbf{Selecting `good' universum samples} The effectiveness of DOC$^3$ also depends on the type of universum used. Our analysis in section \ref{sec_results_MVTEC_corr} provides some initial insights into the workings of DOC$^3$, and how to loosely identify `bad' contradictions. Additional analysis, possibly inline with the Histogram of Projections (HOP) technique introduced in \cite{cherkassky2011practical,dhar2019}, is needed to improve our understanding of `good' universum samples. This is an open research problem.

\end{document}